\pdfoutput=1

\documentclass[11pt]{article}

\usepackage[final]{acl}
\usepackage{times}
\usepackage{latexsym}
\usepackage{float}
\usepackage[T1]{fontenc}
\usepackage[utf8]{inputenc}
\usepackage{microtype}
\usepackage{inconsolata}
\usepackage{graphicx}

\usepackage{algorithm}
\usepackage{listings}
\usepackage{float}
\usepackage{booktabs}
\usepackage{longtable}
\usepackage{algpseudocode}
\usepackage{multirow}
\usepackage{amsmath}
\usepackage{bm}
\usepackage{amssymb}
\usepackage{makecell}
\usepackage{xspace}
\usepackage{textcomp}
\usepackage{subcaption}

\usepackage{xcolor}
\usepackage{colortbl}
\usepackage{tcolorbox}

\newcommand{\mname}{{\sc Tag-Instruct}\xspace}

\title{\mname: Controlled Instruction Complexity Enhancement \\ Through Structure-Based Augmentation}
\author{
  He Zhu$^{1}$, Zhiwen Ruan$^2$, Junyou Su$^{1}$, Xingwei He$^{5}$\\
  \textbf{Yun Chen$^4$, Wenjia Zhang$^{1,3*}$ Guanhua Chen$^2$\thanks{*Co-corresponding authors: chengh3@sustech.edu.cn, wenjiazhang@tongji.edu.cn}} \\
  $^1$ Peking University, 
  $^2$ Southern University of Science and Technology,
  $^3$ Tongji University \\
  $^4$ Shanghai University of Finance and Economics,
  $^5$ The University of Hong Kong \\
}

\begin{document}
\maketitle

\begin{abstract}
High-quality instruction data is crucial for developing large language models (LLMs), yet existing approaches struggle to effectively control instruction complexity. We present \mname, a novel framework that enhances instruction complexity through structured semantic compression and controlled difficulty augmentation. Unlike previous prompt-based methods operating on raw text, \mname compresses instructions into a compact tag space and systematically enhances complexity through RL-guided tag expansion. Through extensive experiments, we show that \mname outperforms existing instruction complexity augmentation approaches. Our analysis reveals that operating in tag space provides superior controllability and stability across different instruction synthesis frameworks.
\end{abstract}

\section{Introduction}

Large language models (LLMs) have demonstrated remarkable success across diverse tasks \citep{gpt4,llama3,qwen25,deepseekv3}, from natural language understanding \citep{phi4} to complex reasoning \citep{deepseek-r1,qwen2}. To fully realize their potential, these models require high-quality instruction-tuning data, which plays a crucial role in model post-training and reinforcement learning initialization \citep{lima,limo,dpo}.  This recognition has led to extensive research in instruction data synthesis \citep{self_instruct,flanv2,magpie}.

\begin{figure}[h]
    \centering
    \includegraphics[width=1\columnwidth]{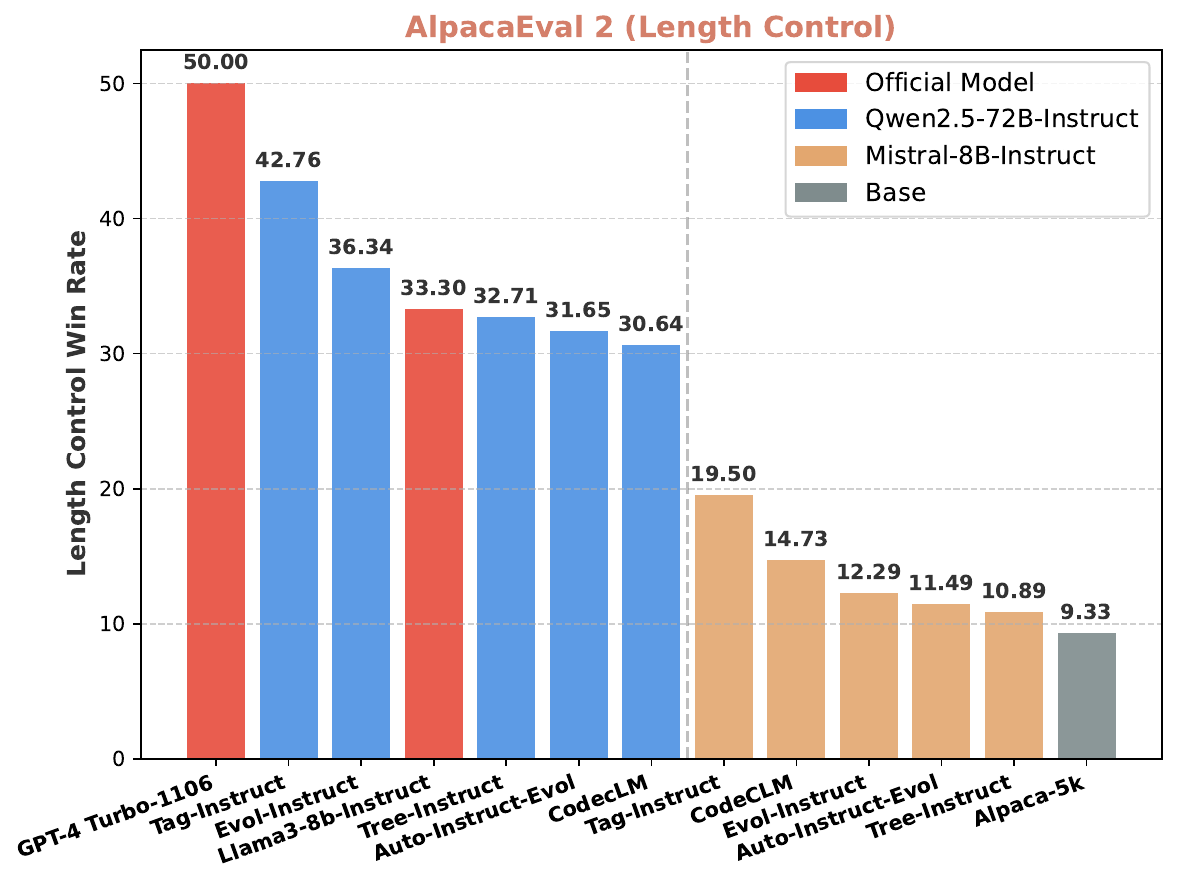} 
    \captionsetup{font=small, labelfont=bf}
    \caption{\textbf{AlpacaEval 2 Length Control Win Rate.} \mname achieves the highest win rate among all compared methods. \textcolor{red}{Red} bars represent official released models. For other methods, all are fine-tuned on LLaMA3-8B base model, where \textcolor{blue}{blue}, \textcolor{orange}{orange}, and \textcolor{gray}{gray} bars represent methods using Qwen2.5-72B-Instruct as teacher model, methods using Mistral-8B-Instruct as teacher model, and base model respectively. Evaluated using \textbf{GPT-4o-Mini}.}
    \label{fig:alpacaeval}
\end{figure}

While recent work has made progress in synthetic data generation \cite{fanno, magpie}, current instruction datasets still lack sufficient complexity to fully develop model capabilities \cite{wizardlm}. Studies show that model performance strongly correlates with instruction difficulty \cite{tree-instruct, zhao2024long}, yet existing approaches struggle to effectively control and scale instruction difficulty. Specifically, \citet{wizardlm} and its follow-up works \cite{codeclm, auto-instruct-evol} rely on models' prior knowledge for self-reflection on augmentation directions, typically using prompts to directly generate ``harder'' instructions. However, this approach lacks precise control over difficulty progression, as the concept of ``harder'' remains ambiguous and unquantifiable, making it challenging to systematically guide the augmentation process. While \citet{tree-instruct} attempts to address this by discretizing instructions into semantic trees for manipulation, it still struggles to identify which semantic components are worth augmenting and how to effectively guide the augmentation direction. We name these approaches as \textit{Prompt-based Augmentation}, which face two fundamental challenges: (1) difficulty in identifying which semantic components are worth augmenting, and (2) lack of principled guidance on which augmentation direction would lead to meaningful complexity increases.

From a linguistic perspective, instructions contain both essential semantic components with high information density and auxiliary content that merely contributes to fluency \cite{kemp2018semantic}. This observation reveals why \textit{Prompt-based Augmentation} struggles - by operating directly on raw instructions, it works in an unnecessarily complex search space that includes both essential and auxiliary content. To address these challenges, we propose \textit{Structure-based Augmentation}, drawing inspiration from representation learning in Variational Autoencoders (VAEs) \citep{kingma2013auto,sentencevae}. Our key insight is that by compressing instructions into a structured tag space, we can (1) identify valuable semantic components by focusing only on essential information, and (2) guide augmentation direction by quantitatively measuring each tag's contribution to instruction complexity. This transforms the augmentation problem from unstructured token-level modifications to systematic concept-level operations \cite{largeconceptmodel}.

Building on this intuition, we propose \mname, a novel \textbf{Compress \& Operation} framework for controlled difficulty augmentation. Our framework operates through a three-stage iterative pipeline: (1) semantic encoding that compresses instructions into a compact tag representation while preserving task-relevant information, (2) RL-based tag expansion that explores the tag space to discover new semantic tags that meaningfully increase complexity. 
(3) instruction synthesis that generates enhanced instructions by conditioning on both the expanded tag set and original instruction. This iterative process enables progressive difficulty increases while maintaining semantic coherence.

Extensive experiments demonstrate the effectiveness of \mname across diverse settings. Using Alpaca-5k \cite{alpaca} as the base dataset, our method enables LLaMA-3-8B to achieve performance comparable to GPT-4-Turbo on standard benchmarks, significantly outperforming existing instruction augmentation approaches. Through detailed ablation studies, we verified that operating in tag space provides superior controllability and makes different instruction synthesis frameworks more stable and effective.\footnote{Our code is publicly available at \url{https://github.com/sustech-nlp/Tag-Instruct}.} 

\begin{figure*}[h]
    \centering
    \includegraphics[width=1\textwidth]{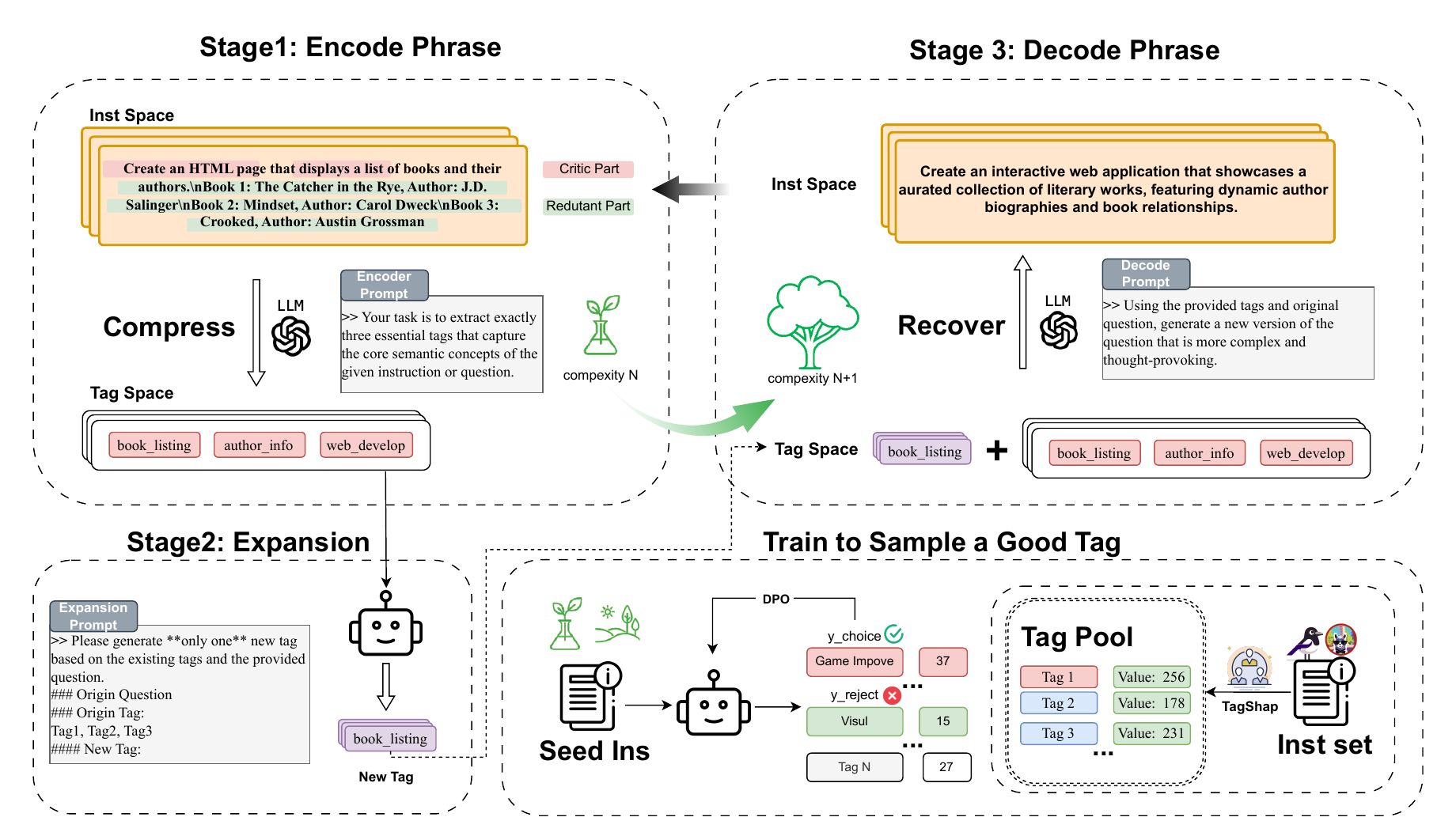} 
    \caption{Overview of \mname framework. The framework operates in three stages: (1) \textbf{Encode Phrase}: Semantic Encoding compresses instructions into tag representations, (2) \textbf{Expansion}: Controlled Tag Expansion explores the tag space to increase complexity while maintaining semantic coherence through iterative difficulty augmentation, and (3) \textbf{Decode Phrase}: Instruction Synthesis generates enhanced instructions from the expanded tag set. The iterative process allows for progressive difficulty increases while preserving semantic validity.}
    \label{fig:tag-instruct}
\end{figure*}

\section{Related Work}
\paragraph{Instruction Data Augmentation} \label{related:dataaug} High-quality instruction data is fundamental for model alignment and performance \cite{llama3,nemotron4340btechnicalreport}. For instruction synthesis from scratch, \citet{self_instruct} proposed automatic generation through bootstrapping from human demonstrations, though constrained by seed data. \citet{magpie} advanced this by introducing pre-query templates for direct instruction construction, while \citet{fanno} improved data quality using tagging-based prompts and UCB-based bootstrapping. \citet{1mperson} introduced a persona-driven approach using 1 billion diverse personas from web data to create synthetic instructions at scale. For instruction enhancement based on existing data, several methods focus on difficulty progression: \citet{wizardlm} and \citet{auto-instruct-evol} explored evolutionary approaches for iterative complexity increase, \citet{tree-instruct} proposed semantic tree structures for controlled augmentation, and \citet{codeclm} leveraged encode-decode principles. However, existing enhancement methods still lack precise control (especially numerical parameterization) over difficulty progression, motivating our structured tag-based approach.

\paragraph{Information Compression in Hidden Space} Drawing inspiration from variational autoencoders~\citep{vae} which learn disentangled latent representations, researchers have investigated diverse approaches to compress instructions into abstract spaces to enhance semantic understanding. One prominent direction focuses on morphological abstraction, where instructions are systematically decomposed into skill-oriented tags and hierarchical concepts~\citep{instag,metacognitive,instructskillmix,codeclm,largeconceptmodel}. Another stream of research explores functional compression by transforming complex problem-solving procedures into executable representations, such as programmatic structures or formal planning schemas~\citep{stepback, plansearch, distilled}. Complementing these approaches, researchers have also investigated continuous latent space mappings of symbolic instructions to capture underlying semantic regularities and logical relationships~\citep{sentencevae,coconut}. The demonstrated efficacy of these compression-based abstraction techniques motivates our work to leverage structured latent representations for controlled instruction generation. Recent information compression works like \citet{instructskillmix} and 
\citet{metacognitive} focus on instruction synthesis from scratch, which is orthogonal to our controlled difficulty augmentation approach. While prior work~\cite{codeclm} directly decodes from compressed metadata after self-observation (Self-Rubrics and action sampling), our approach uniquely operates in tag space for tag complexity enhancement before instruction reconstruction, enabling more controlled instruction generation.

\paragraph{Data Quality Investigation} Recent works have extensively explored what constitutes high-quality instruction data for large language models. Early approaches by \citet{liu2023what} and \citet{du2023mods} utilize fine-tuned models and specialized open-source LLMs for data quality assessment. Following works by \citet{instruct-mining} and \citet{wettig2024qurating} propose automatic metrics to evaluate data quality through natural language indicators. Several studies by \citet{zhao2024long} reveal a surprisingly simple yet effective finding that longer responses often contain more learnable information. Other approaches by \citet{LESS} and \citet{cherryllm} introduce optimizer-aware selection and instruction difficulty metrics to identify high-quality samples. Recent work by \citet{self-distillation} further emphasizes the importance of distribution alignment between fine-tuning data and the model's original capabilities. These studies collectively indicate that high-quality instruction data should be difficult, human-preference-aligned, and close to the pre-trained model's distribution -- properties that often manifest in longer responses. This observation inspires our measurement of instruction utility.

\section{Method} \label{sec:method}
\subsection{Motivation and Overview} \label{sec:motivation}

The quality of synthesized data correlates to correctness, diversity and complexity \cite{liu2023deita,fanno}. While recent works \cite{self_instruct,kaur2025instructskillmix} primarily focus on enhancing instruction diversity, we systematically address instruction complexity enhancement, aligning with previous approaches \cite{wizardlm,tree-instruct}. Following the standard assumption in data synthesis works of having access to a capable instruction-following teacher model, we propose \mname for controlled complexity augmentation.

Existing prompt-based augmentation methods \cite{wizardlm,tree-instruct} operate directly in token space, lacking quantitative control mechanisms and principled guidance for complexity enhancement. To overcome these limitations, \mname operates through an \textbf{Encode → Expansion → Decode} paradigm that fundamentally differs from existing token-space methods by compressing instructions into a structured tag space for systematic concept-level operations. For each input instruction $x$, our framework operates through three stages: \textbf{Stage (1) Encode}: compresses instruction $x$ into semantic tags $z_{\text{base}}$ (Section~\ref{sec:semantic_encoding}); \textbf{Stage (2) Expansion}: expands the tag set using RL-guided selection to identify high-utility tags $z_{\text{new}}$, forming $z_{\text{base}'} = z_{\text{new}} \cup z_{\text{base}}$ (Section~\ref{sec:controlled_tag_expansion}) guided by Shapley value estimation (Section~\ref{sec:utility_estimation}); \textbf{Stage (3) Decode}: reconstructs an enhanced instruction $x'$ from the expanded tags and original instruction $\{z_{\text{base}}', x\} \rightarrow x'$ (Section~\ref{sec:ins_reconstruct}). This approach is orthogonal to diversity-focused methods and can be integrated with existing synthesis frameworks. Please refer to Figure~\ref{fig:tag-instruct} for the framework overview.

\subsection{Semantic Compression} \label{sec:semantic_encoding}
Inspired by the ideas of variational representation learning and information bottleneck principle \cite{kingma2013auto,sentencevae}, \mname optimizes the data complexity in the compressed discrete tag space. The tag space exhibits reduced search complexity while preserving task-relevant information. Thereby this semantic compression helps to filter out extraneous elements from the instructions and facilitates the tag expansion, resulting in more effective complexity enhancement.

Given an instruction $x$ from existing synthesized data, the teacher LLM is prompted (see the instruction-to-tag prompt in Appendix~\ref{appendix:encode_prompt}) to generate relevant tags $z_{\mathrm{base}}$ of the input instruction. The crafted prompt is expected to explicitly guide the model to extract tags that preserve core semantic meanings with the input. Among these tags, those related to action-oriented concepts and task objectives are emphasized and with higher priority as they are indicative of the input intension. For example, the instruction \textit{Create an HTML page that displays a list of books and their authors} is compressed into \{\textit{book\_listing}, \textit{author\_info}, \textit{web\_develop}\}. We discuss and compare different prompt guidelines for semantic compression in Section~\ref{exp:encode_prompt_selection}.

\subsection{Tag Complexity Expansion} \label{sec:controlled_tag_expansion}

The compressed tags $z_{\mathrm{base}}$ are further expanded for complexity optimization. The potential invited tags $z_{\mathrm{new}}$ are determined based on the following criteria. 
(1) \textbf{Tag Utility} ($R_{\text{utility}}(z_{\text{new}})$): Measures how much a new tag contributes to the instruction complexity. For example, adding \textit{responsive\_design} to our HTML example introduces meaningful complexity by requiring additional technical considerations, while \textit{simple\_styling} adds little value. 
(2) \textbf{Semantic Alignment} ($R_{\text{align}}(z_{\text{new}}, z_{\text{base}})$): Ensures compatibility between new and existing tags. For instance, \textit{data\_analysis} might be a valuable tag but conflicts with our web development context, whereas \textit{user\_authentication} aligns naturally with the existing web application tags. 
(3) \textbf{Global Coherence} ($R_{\text{coherence}}(z_{\text{new}} \cup z_{\text{base}}, x)$): Verifies that the expanded tag set can generate executable instructions. Tags like \textit{database\_integration} and \textit{web\_develop} can naturally combine into coherent tasks, while \textit{quantum\_computing} and \textit{web\_develop} might create unrealistic requirements.


While prompt-based methods can reasonably assess semantic alignment and global coherence through conceptual compatibility and feasibility checks that suit LLM strengths, tag utility quantification requires more sophisticated approaches. Although one could directly prompt the LLM to evaluate tag utility \cite{zheng2023judgingllmasajudgemtbenchchatbot}, this method inadequately models complex tag interactions and provides no principled evaluation framework. \mname is proposed to incorporate a policy model trained with preference data on tag expansion task to generate augmented tags with high utility.
The preference data are collected by tag utility quantification with Shapley value estimation \cite{shapley1953value,tokenshap} (detailed in Section~\ref{sec:utility_estimation}). The tags with high utility are selected as positive while the opposite are set as rejected samples. Initialized from the teacher LLM, the policy model is then optimized and adapted to the tag expansion task with alignment tuning methods like DPO \cite{dpo}. 
During the tag complexity extension stage, the expanded tags are generated by prompting the optimized policy model with the detailed requirements of semantic alignment and global coherence criteria.

\subsection{Utility Estimation through Shapley Values} \label{sec:utility_estimation}

As discussed in Section \ref{sec:controlled_tag_expansion}, quantification of tag utility presents a unique challenge that requires a more principled approach. Inspired by cooperative game theory \cite{shapley1953value,tokenshap}, we propose to estimate the tag utility by measuring their marginal contributions in instruction construction with Shapley value \cite{shapley1953value,tokenshap}.

We frame instruction complexity enhancement as a cooperative game where semantic tags $z_i \in Z$ are players collaborating to construct effective instructions. Each tag $z_i$ participates in multiple instruction construction games by combining with different tag subsets $S \subseteq Z$. The core insight is that a tag's true utility $\phi_i$ can be measured through its marginal contribution $v(S \cup \{z_i\}) - v(S)$ to instruction quality across all possible collaborations $S$, capturing complex interaction effects in the compressed representation space $\mathcal{Z}$. 
Formally, we calculate each tag's Shapley value as $\hat{\phi}_i = \frac{1}{N}\sum_{j=1}^N [v(S_j \cup \{z_i\}) - v(S_j)]$, where $N$ is the number of instruction construction games. However, the computation is intractable due to the vast number of potential tag subsets $S$.

We simplify the computation process and estimate the Shapley value based on an existing instruction-response dataset $\mathcal{D}^t = {(x_i, y_i)}_{i=1}^M$ such as Alpaca-Cleaned \cite{alpaca} and Tulu-mixture \cite{tulu3}. This dataset is different from the instruction data which is to be optimized for better complexity with \mname. 
The tags are extracted from the instructions following the practice in Section~\ref{sec:semantic_encoding}. Following the suggestions in \citet{zhao2024long} and \citet{shen2024rethinkingdataselectionsupervised}, we use response length as a proxy for instruction quality $v$, as it effectively captures both complexity and information density.
To address the computational complexity of Shapley value calculation, we approximate the tag utility as the average response length of instructions containing that tag: 
\begin{equation} \label{eq:reward}
    \phi_i = \frac{\sum_{j \in \mathcal{D}^t_i} |y_j|}{|\mathcal{D}^t_i|},
\end{equation}
where $\mathcal{D}^t_i$ denotes instructions containing tag $i$ and $|y_j|$ is the token length of response $j$. 
With the dataset $\mathcal{D}^t$, the reward of each tag can be then computed with Equation~\ref{eq:reward}. Please refer to Appendix~\ref{appendix:rl_expansion} for more details about how to collect the preference data for optimizing the policy model on the tag expansion task.

\subsection{Instruction Reconstruction} \label{sec:ins_reconstruct}

After determining the augmented tags, \mname then reconstruct the instruction based on the original instruction, base tags as well as augmented tags by prompting the teacher LLM.

The augmented instruction is optimized iteratively, where the output instruction of the $(i-1)^\mathrm{th}$ iteration is the input of the $i^\mathrm{th}$ iteration with \mname. Through this iterative process, each subsequent iteration introduces additional complexity and challenges, as new tags and requirements are incorporated to progressively increase the difficulty level of the instruction.

\begin{table*}[tbhp]
    \centering
    \vspace{1em}
    \resizebox{0.9\textwidth}{!}{%
    \begin{tabular}{l c c c c c c}
        \toprule
        \multirow{2}{*}{\textbf{Setup}} & \multirow{2}{*}{\textbf{\#Convs}} & \multicolumn{3}{c}{\textbf{AlpacaEval 2.0}} & \textbf{Arena-Hard} & \textbf{MT-Bench} \\
        \cmidrule(lr){3-5} \cmidrule(lr){6-6} \cmidrule(lr){7-7}
        & & LC (\%) & WR (\%) & SD & Score (95\% CI) & Score \\
        \midrule
        \multicolumn{7}{l}{\textbf{Base Model: LLaMA3-8b / Teacher Model: Ministral-Instruct-8b}} \\
        \midrule
        Alpaca-5k (baseline) & 5K & 9.33 & 5.09 & 0.74 & 3.1 (-0.7, 0.7) & 5.20 \\
        \midrule
        + Evol-Instruct & 5K & 12.29 {\color{olive}(+2.96)} & 9.90 {\color{olive}(+4.81)} & 0.99 & 16.8 (-1.7, 1.6) {\color{olive}(+13.7)} & 5.68 {\color{olive}(+0.48)} \\
        + Auto-Instruct-Evol & 5K & 11.49 {\color{olive}(+2.16)} & 10.19 {\color{olive}(+5.10)} & 1.01 & 16.8 (-1.8, 1.4) {\color{olive}(+13.7)} & 5.89 {\color{olive}(+0.69)} \\
        + Tree-Instruct & 5K & 10.89 {\color{olive}(+1.56)} & 10.47 {\color{olive}(+5.38)} & 1.01 & 16.6 (-1.6, 1.5) {\color{olive}(+13.5)} & 5.94 {\color{olive}(+0.74)} \\
        + CodecLM & 5K & 14.73 {\color{olive}(+5.40)} & 12.71 {\color{olive}(+7.62)} & 1.10 & 19.2 (-1.4, 1.5) {\color{olive}(+16.1)} & 6.00 {\color{olive}(+0.80)} \\
        \midrule
        Alpaca-Clean & 52K & 7.73 {\color{red}(-1.60)} & 5.21 {\color{olive}(+0.12)} & 0.74 & 3.0 (-0.4, 0.5) {\color{red}(-0.1)} & 4.89 {\color{red}(-0.31)} \\
        WizardLM-data & 192K & 11.68 {\color{olive}(+2.35)} & 5.69 {\color{olive}(+0.60)} & 0.75 & 4.5 (-0.8, 1.1) {\color{olive}(+1.4)} & 5.36 {\color{olive}(+0.16)} \\
        \rowcolor{blue!10} \mname & 5K & \textbf{19.50} {\color{olive}(+10.17)} & \textbf{19.21} {\color{olive}(+14.12)} & 1.30 & \textbf{22.1} (-2.1, 1.9) {\color{olive}(+19.0)} & \textbf{6.15} {\color{olive}(+0.95)} \\ 
        \midrule \midrule
        \multicolumn{7}{l}{\textbf{Base Model: LLaMA3.2-3b / Teacher Model: Ministral-Instruct-8b}} \\
        \midrule
        Alpaca-5k (baseline) & 5K & 8.52 & 4.73 & 0.71 & 2.6 (-0.5, 0.7) & 4.21 \\
        \midrule
        + Evol-Instruct & 5K & 9.06 {\color{olive}(+0.54)} & 7.48 {\color{olive}(+2.75)} & 0.85 & 12.7 (-1.5, 1.4) {\color{olive}(+10.1)} & 4.95 {\color{olive}(+0.74)} \\
        + Auto-Instruct-Evol & 5K & 8.79 {\color{olive}(+0.27)} & 8.79 {\color{olive}(+4.06)} & 0.95 & 10.8 (-1.2, 1.2) {\color{olive}(+8.2)} & 5.18 {\color{olive}(+0.97)} \\
        + Tree-Instruct & 5K & 10.01 {\color{olive}(+1.49)} & 8.50 {\color{olive}(+3.77)} & 0.91 & 10.4 (-1.2, 1.4) {\color{olive}(+7.8)} & 5.18 {\color{olive}(+0.97)} \\
        + CodecLM & 5K & 11.24 {\color{olive}(+2.72)} & 11.19 {\color{olive}(+6.46)} & 1.03 & 12.7 (-1.1, 1.2) {\color{olive}(+10.1)} & 4.79 {\color{olive}(+0.58)} \\
        \midrule
        Alpaca-Clean & 52K & 8.06 {\color{red}(-0.46)} & 4.12 {\color{red}(-0.61)} & 0.67 & 2.8 (-0.6, 0.8) {\color{olive}(+0.2)} & 4.74 {\color{olive}(+0.53)} \\
        WizardLM-data & 192K & 6.74 {\color{red}(-1.78)} & 4.46 {\color{red}(-0.27)} & 0.68 & 3.7 (-0.6, 0.9) {\color{olive}(+1.1)} & 4.94 {\color{olive}(+0.73)} \\
        \rowcolor{blue!10} \mname & 5K & \textbf{14.28} {\color{olive}(+5.76)} & \textbf{14.81} {\color{olive}(+10.08)} & 1.19 & \textbf{12.9} (-1.2, 1.1) {\color{olive}(+10.3)} & \textbf{5.21} {\color{olive}(+1.00)} \\
        \bottomrule
    \end{tabular}}
    \caption{Performance comparison of instruction-tuned models on LLaMA3-8b and LLaMA3.2-3b base models. We report Length Control (LC) and Win Rate (WR) from AlpacaEval 2.0, Standard Deviation (SD) of model outputs, and Arena-Hard scores with 95\% Confidence Intervals (CI). \textbf{Bold} numbers indicate best performance across all metrics, while rows with \colorbox{blue!10}{blue background} highlight our \mname approach.}
    \label{tab:use_tag}
\end{table*}

\section{Experiments}

\subsection{Setup}

\paragraph{Data and Model Settings}
We conducted experiments using Alpaca-Clean dataset~\citep{alpaca}, from which we randomly sampled 5K instructions to create Alpaca-5k as our initial instruction set. And we use LLaMA-3-8B and LLaMA-3.2-3B as our base models, and Ministral-Instruct-8b~\citep{jiang2023mistral7b} as the teacher model. For data pool construction ~\ref{sec:utility_estimation}, we extract tags from instructions following \citet{instag} and utilize instructions from \citet{magpie} to estimate tag rewards and train our policy model for tag expansion. Detailed specifications of the datasets, models, and training procedures are provided in Appendix~\ref{appendix:exp_setup}.

\paragraph{Evaluation Benchmarks}
We compared \mname with baselines on the following benchmarks: (1) \textbf{AlpacaEval 2.0}~\citep{alpaca_eval,lengthcontrolledalpaca} is an automated evaluation framework based on a annotation model(GPT-4). By comparing responses generated by two different models for the same set of 805 prompts, AlpacaEval computes the pairwise win rate, automating the evaluation process. (2) \textbf{MT-Bench}~\citep{zheng2024judging} is aimed at assessing the conversational and instruction-following abilities of LLMs. A one-shot chat template is used to test all models in our experiments. (3) \textbf{ArenaHard-Auto}~\citep{arenahard} is an automatic evaluation tool for instruction-tuned LLMs. It includes 500 challenging queries sourced from Chatbot Arena, evaluated against a baseline model (GPT-4-0314). For all evaluations, we used GPT-4o-mini as the judge model given our limited budget. 

\paragraph{Baselines}
We evaluated our method against several state-of-the-art instruction augmentation methods and popular instruction-tuning datasets. The methodological baselines included Self-Instruct~\citep{self_instruct}, Evol-Instruct~\citep{wizardlm}, Tree-Instruct~\citep{tree-instruct}, Auto-Instruct-Evol~\citep{auto-instruct-evol}, and CodecLM~\citep{codeclm}. For data baselines, we compared against Alpaca-Clean~\citep{alpaca} and WizardLM-data~\citep{wizardlm}. Detailed descriptions of each baseline can be found in Appendix~\ref{appendix:baselines}.

\subsection{Results} \label{sec:main_results}
We present experimental results in Table~\ref{tab:use_tag}, demonstrating the superior performance of our \mname approach across model scales and evaluation metrics. On the LLaMA3-8b model, \mname shows significant improvements over the Alpaca-5k baseline: +10.17\% in Length Control and +14.12\% in Win Rate on AlpacaEval 2.0, +19.0 points on Arena-Hard, and +0.95 points on MT-Bench. Notably, \mname achieves this using only 5K conversations, outperforming larger datasets like WizardLM (192K conversations) which shows only 2.35\% Length Control and 0.60\% Win Rate improvements. On LLaMA3.2-3b, our method shows strong performance with +5.76\% Length Control and +10.08\% Win Rate improvements over baseline, exceeding CodecLM (+2.72\% and +6.46\%). Larger datasets like Alpaca-Clean (52K) and WizardLM (192K) show decreased performance on the 3b model, highlighting our approach's value for smaller architectures. Across all metrics, \mname outperforms existing methods including CodecLM. On the 8b model, we achieve +4.77\% Length Control and +6.50\% Win Rate improvements, while on the 3b model, we show +3.04\% Length Control and +3.62\% Win Rate gains. These improvements demonstrate the effectiveness of our tag-based approach.

\subsection{Ablation Study}

\begin{table}[t]
    \centering
    \begin{tabular}{l@{\hspace{6pt}}r@{\hspace{6pt}}r@{\hspace{6pt}}r@{\hspace{6pt}}r}
        \toprule
        \textbf{Strategy} & \textbf{\#Inst} & \textbf{\#Resp} & \textbf{AE2.0} & \textbf{AH} \\
        \midrule

        Base (Alpaca-5k) & 21.0 & 159.5 & 8.78 & 3.1 \\
        Prompt-based & 287.3 & 1006.7 & 15.43 & 19.1 \\
        \rowcolor{blue!10} RL-based & 285.8 & \textbf{1037.0} & \textbf{19.50} & \textbf{22.1} \\
        \bottomrule
    \end{tabular}
    \caption{Performance comparison of different tag expansion strategies. \#Inst represents instruction length, \#Resp represents response length, AE2.0 represents AlpacaEval2.0 length-controlled win rate, AH represents ArenaHard win rate.}
    \label{tab:tag_sampling}
\end{table}

\paragraph{Impact of Reward Model Guidance.}
As discussed in Section~\ref{sec:controlled_tag_expansion}, we compare different approaches to expand new tags given the input instruction and base tags. The vanilla prompt-based method directly generates new tags with crafted prompt (see in Appendix~\ref{appendix:adding_tag_template}). The RL-based approach is incorporated in \mname which uses optimized policy model for generation of tags with high utility.The results in Table~\ref{tab:tag_sampling} show that RL-based sampling outperformed random sampling in the tag expansion task. With comparable instruction lengths (285.8 vs 287.3 tokens), the RL-guided approach generated longer responses (1037.0 vs 1006.7 tokens) and achieved better performance on Alpaca-Eval 2.0 (19.50\% vs 15.43\%) and ArenaHard (22.1\% vs 19.1\%). This improvement demonstrates that our policy model effectively encourages the generation of high-quality tags which can improve the data complexity.

\begin{figure*}[htbp]
    \centering
    \begin{subfigure}[b]{0.32\textwidth}
        \includegraphics[width=\textwidth]{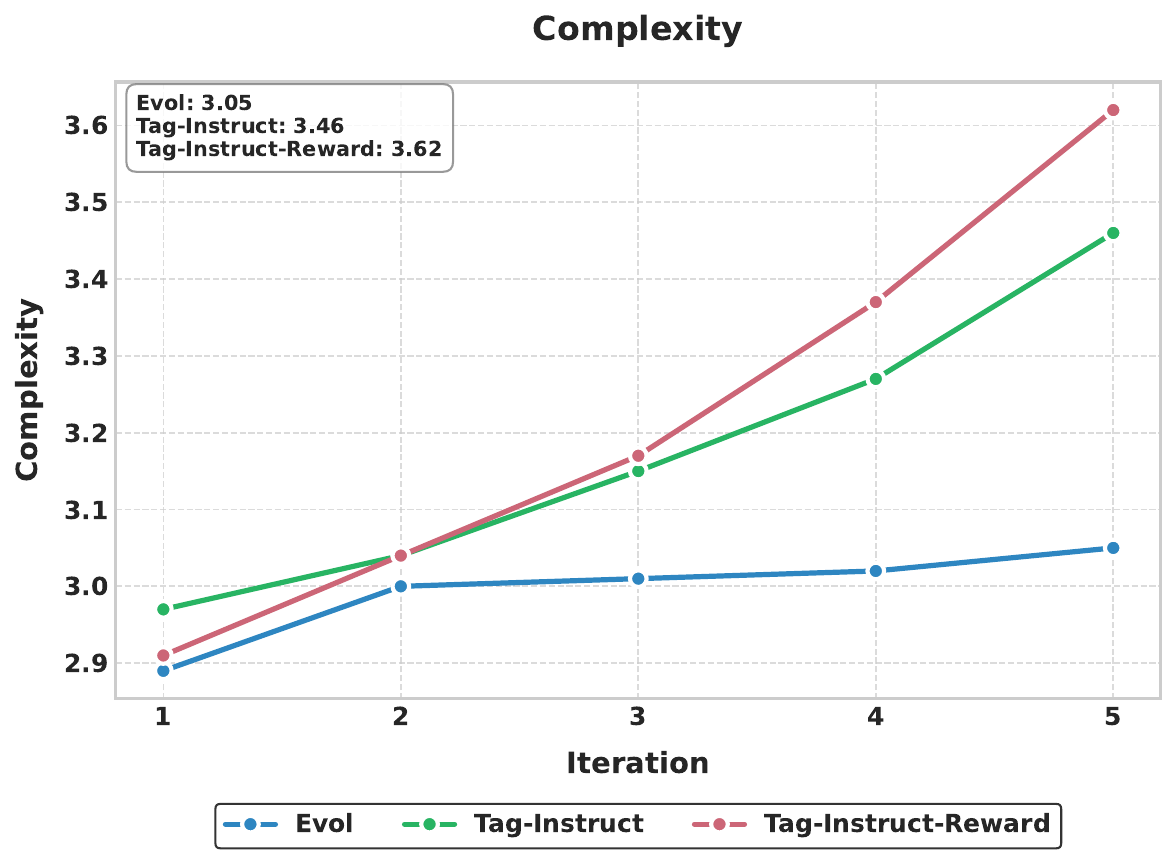}
        \caption{Complexity Analysis}
        \label{fig:iteration_complexity}
    \end{subfigure}
    \begin{subfigure}[b]{0.32\textwidth}
        \includegraphics[width=\textwidth]{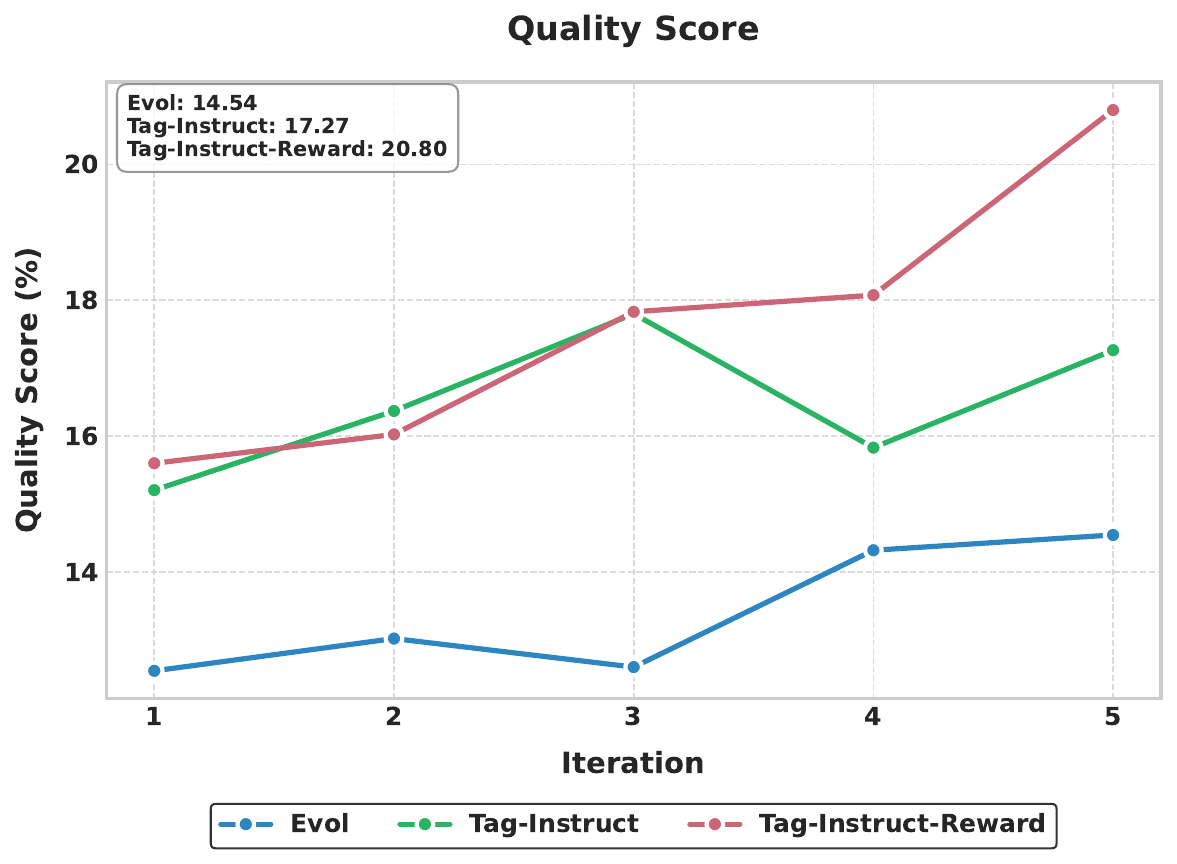}
        \caption{Quality Analysis}
        \label{fig:iteration_quality}
    \end{subfigure}
    \begin{subfigure}[b]{0.32\textwidth}
        \includegraphics[width=\textwidth]{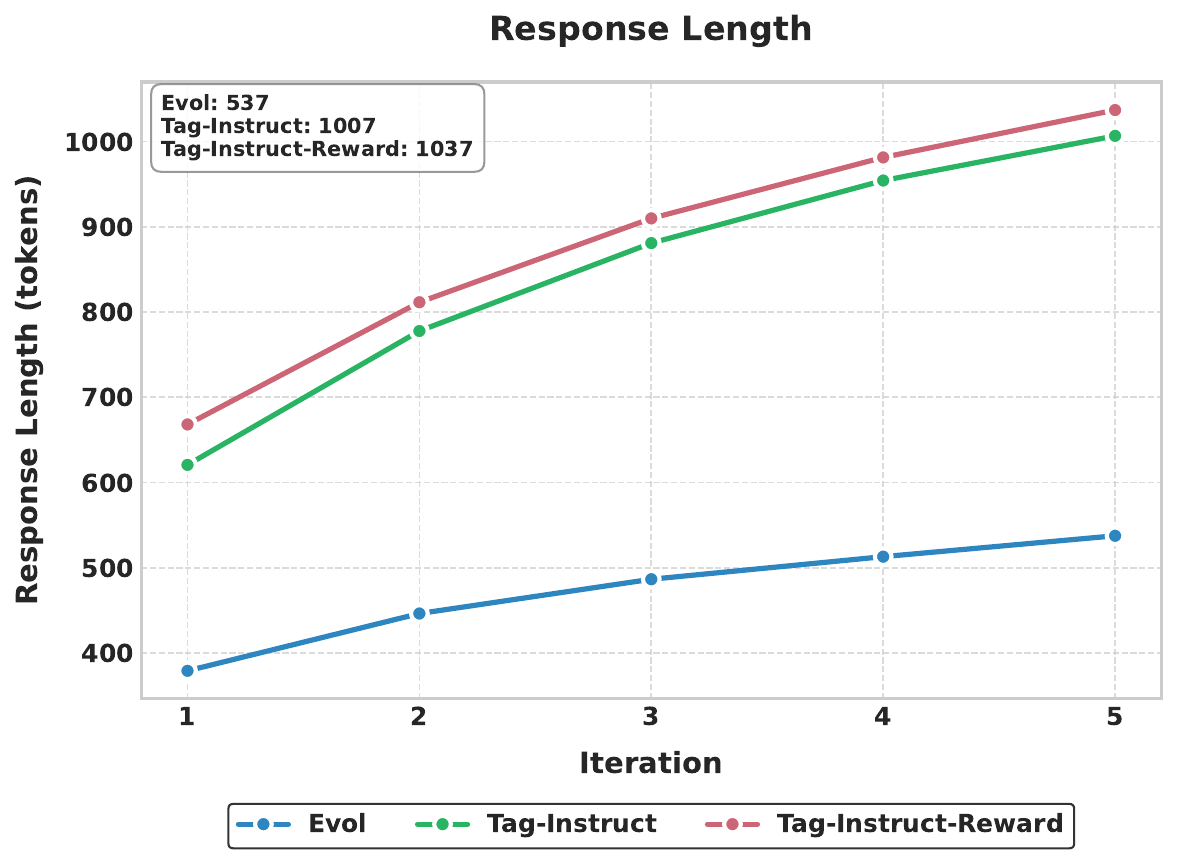}
        \caption{Response Length Analysis}
        \label{fig:iteration_response}
    \end{subfigure}
    \caption{\textbf{Iterative analysis of instruction generation.} Comparison between Evol-Instruct (baseline), \mname (prompt-based), and \mname (RL-based) across: (a) Instruction complexity via Instagger tags; (b) Quality score (average of Arena-Hard and Alpaca-Eval 2.0); (c) Response length.}
    \label{fig:iteration_analysis}
\end{figure*}

\paragraph{Evaluation of Iterative Optimization.}
To assess the effectiveness of our iterative framework, we analyzed the progression across five iterations, as shown in Figure~\ref{fig:iteration_analysis}. We compared three approaches: Evol-Instruct (baseline), \mname (using prompt-based tag expansion), and \mname-Reward (using RL-trained tag generator). Following established frameworks\cite{magpie}, we used the average of Arena-Hard and Alpaca-Eval 2.0 scores for quality assessment, and Instagger\cite{instag} tag counts for complexity measurement. As shown in Figure~\ref{fig:iteration_complexity}, \mname-Reward consistently achieves higher complexity scores across iterations compared to both \mname and Evol baselines. The quality metrics in Figure~\ref{fig:iteration_quality} demonstrate a similar pattern, with \mname-Reward showing steeper improvement and reaching higher final performance. Analysis of response lengths (Figure~\ref{fig:iteration_response}) further validates the effectiveness of our approach, with \mname variants generating substantially longer responses than Evol. The superior performance of \mname-Reward over prompt-based \mname confirms that our RL-based tag generator enables more controlled and effective instruction enhancement through structured tag-space operations.

\section{Analyses}

\subsection{Different Crafted Prompts for Semantic Compression} \label{exp:encode_prompt_selection}

\begin{table*}[t]
    \centering
    \resizebox{\textwidth}{!}{
    \begin{tabular}{l|cc|cc|c}
        \toprule
        \textbf{Method} & \textbf{AlpacaEval2.0 LC} & \textbf{ArenaHard} & \textbf{Similarity} & \textbf{ROUGE-L} & \textbf{Intent Tag} \\
        \midrule
        Basic     & 12.9182 & 16.1 (-1.7, 1.5) & 0.6536 & 0.3113 & False \\
        Enhanced  & 13.1831 & 16.7 (-1.2, 1.6) & 0.6023 & 0.2778 & True \\
        Evolved   & 14.1889 & \textbf{18.0 (-1.8, 1.4)} & 0.5373 & 0.1992 & True \\
        \midrule
        Model-based\cite{instag} & \textbf{14.3498} & 17.6 (-1.8, 2.1) & 0.3600 & 0.1667 & True \\
        \bottomrule
    \end{tabular}}
    \caption{Analysis of prompt evolution steps and their impact on reconstruction quality and downstream performance.}
    \label{tab:tag_extraction}
 \end{table*}
 
 A critical challenge in instruction augmentation is achieving meaningful complexity enhancement while preventing semantic drift from the original instruction. To address this, we compare four prompt strategies for semantic compression, each designed with different priorities. We selected 1,000 instructions from Alpaca-5k and conducted five iterations of complexity enhancement to evaluate four encoding strategies: Basic (simple template), Enhanced (joint intention and semantic compression), Evolved (intention-focused encoding), and Model-based (pure intention encoding) \cite{instag}. Detailed prompts for each strategy are provided in Appendix~\ref{appendix:encode_prompt}.
 
 We assessed their performance along two key dimensions: (1) semantic preservation, measured by semantic similarity and ROUGE-L scores between original and reconstructed instructions, and (2) intent extraction capability, which evaluates whether the prompt explicitly requires extracting task intention from instructions.
 
 As shown in Table~\ref{tab:tag_extraction}, a clear pattern emerges: methods with intent extraction capability (Intent Tag = True) consistently outperform the basic approach, with performance gains inversely correlated with semantic similarity. This reveals a fundamental insight—\textit{task-aware compression} that prioritizes functional understanding over surface-level similarity yields superior complexity enhancement. The model-based approach achieves the highest AlpacaEval score (14.3498) despite the lowest similarity (0.3600), demonstrating that aggressive semantic abstraction can be beneficial. However, to maintain instruction coherence and prevent excessive semantic drift during iterative augmentation, we adopt the Evolved strategy, which achieves the best ArenaHard performance (18.0) while preserving moderate semantic alignment (0.5373). This strategic trade-off ensures our framework generates complex yet semantically grounded instructions throughout the iterative enhancement process.

\subsection{Tag Utility Analysis} \label{experiment:tag_analysis}

To better understand the differences between tags with varying utility values, we categorized the tags into quartiles (Q1-Q4) based on their utility values in ascending order. We then analyzed the number of \textbf{derived meanings} for the tags within each quartile. The \textbf{derived meanings} measure captures how many distinct semantic interpretations a tag can reasonably support across different contexts. For instance, while \textit{database\_backup} typically refers to a single concept of data preservation, \textit{security} can encompass authentication mechanisms, encryption protocols, access control systems, and threat detection frameworks. We employed the prompting detailed in Appendix~\ref{appendix:derived_probe}, which is designed according to research on semantic information measurement \cite{semanticentropy}. This prompt systematically probes for multiple meanings before merging similar concepts to form distinct semantic categories.

\begin{figure}[ht]
    \centering
    \includegraphics[width=1\linewidth]{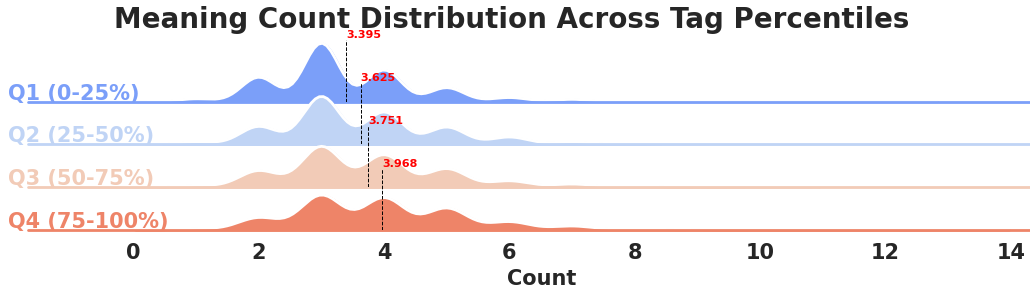}
    \caption{Number of derived meanings across tag utility quartiles (Q1-Q4). Higher utility tags (Q4) exhibited more derived meanings compared to lower utility tags (Q1), suggesting richer semantic content enabled more effective instruction generation. The \textcolor{red}{red} number indicates the mean value.}
    \label{fig:tag_semantic}
\end{figure}

As shown in Figure \ref{fig:tag_semantic}, tags with higher utility values exhibited consistently more derived meanings, indicating they encoded richer semantic information that could be interpreted in more diverse ways. Tags in Q4 (75-100\%) demonstrated the highest semantic complexity with broader meaning distributions, while Q1 tags (0-25\%) showed more constrained semantic interpretations. This semantic richness enabled diverse instruction generation pathways. A comprehensive analysis of additional representative examples and their semantic interpretations is provided in Appendix~\ref{appendix:tag_examples}. This finding suggested that the effectiveness of high-utility tags in generating complex and diverse instructions may be attributed to their inherently \textbf{richer semantic content}.

\subsection{Quantitative Control Through High-Utility Tags}

We investigate whether our utility-based tag selection provides quantitative control beyond expansion operations. \textbf{Tag combination}, which merges distinct semantic concepts to create diverse instructions, is fundamental to recent synthesis methods~\cite{instructskillmix,metacognitive}, yet existing approaches lack principled guidance for selection. Our framework transforms this from heuristic sampling to quantitative optimization through Shapley-based utility scores.

To validate this quantitative advantage, we constructed three tag pools of 1000 tags each from Section~\ref{sec:utility_estimation} using different selection criteria: High-utility (selecting tags with highest Shapley values), Low-utility (selecting tags with lowest Shapley values), and Random (uniformly sampling tags). For each pool, we randomly sampled two tags to combine and decoded them into instructions using the prompting template in Appendix~\ref{appendix:tag_combination}, generating 5,000 instructions per pool. We then fine-tuned models using these instruction sets under identical configurations to ensure fair comparison.

Table~\ref{tab:tag_combination_control} demonstrates the impact of quantitative tag selection. High-utility tags yield 12.3\% longer responses (967.52 vs 861.38 tokens) and achieve 36.1\% relative improvement on AlpacaEval 2.0 (20.21\% vs 14.85\%) compared to random selection—despite generating more concise instructions (94.21 vs 99.09 tokens). This efficiency gain (better performance with shorter prompts) validates that our utility scores capture intrinsic semantic value rather than superficial verbosity. Furthermore, the monotonic relationship between tag utility and downstream performance (Low: 11.02\% → Random: 14.85\% → High: 20.21\% on AlpacaEval) establishes that our framework provides \textbf{quantitative control} over instruction quality—a capability absent in existing prompt-based methods. Ultimately, this demonstrates that structured tag-space operations with utility guidance offer principled mechanisms for instruction synthesis, enabling precise optimization rather than stochastic exploration.

\begin{table}[t]
    \centering
    \begin{tabular}{l@{\hspace{8pt}}r@{\hspace{8pt}}r@{\hspace{8pt}}r@{\hspace{8pt}}r}
        \toprule
        \textbf{Strategy} & \textbf{\#Inst} & \textbf{\#Resp} & \textbf{AE2.0} & \textbf{AH} \\
        \midrule
        Low-utility & 85.02 & 524.87 & 11.02 & 15.8 \\
        Random & 99.09 & 861.38 & 14.85 & 20.9 \\
        High-utility & 94.21 & \textbf{967.52} & \textbf{20.21} & \textbf{24.6} \\
        \bottomrule
    \end{tabular}
    \caption{Quantitative impact of tag utility on synthesis quality. High-utility tags consistently outperform random selection across metrics, validating our Shapley-based scoring mechanism.}
    \label{tab:tag_combination_control}
\end{table}

\subsection{The Impact of Different Teacher Model}
In this section, we explore the performance of \mname with different teacher models. Using the same experimental setup as our main experiments, we compare our approach with baselines by finetuning LLama3-8B using Qwen2.5-72B-instruct \citep{qwen25} as the teacher model. As shown in Table~\ref{tab:teacher_model}, \mname consistently outperforms the baselines, achieving the highest scores on both AlpacaEval2.0 length-controlled tasks (42.76\%) and ArenaHard evaluation (41.2\%). 
These results underscore the broad applicability of our method, surpassing the baseline models across different teacher models. When compared to using Ministral-Instruct-8b as the teacher, Qwen2.5-72B-instruct enhances performance by 23.26 on AlpacaEval2.0 and 19.1 on ArenaHard. This significant improvement can be attributed to two key factors: First, stronger models typically exhibit superior instruction-following capabilities, resulting in higher-quality instruction generation with fewer inconsistencies or unsolvable problems. Second, more capable teacher models possess superior reasoning abilities, allowing them to generate higher-quality responses with more sophisticated analytical thinking and deeper insights. This suggests that leveraging more advanced teacher models can substantially enhance both the complexity and quality of the generated instruction dataset.

\begin{table}[t]
    \centering
    \resizebox{1\columnwidth}{!}{%
    \begin{tabular}{l@{\hspace{8pt}}r@{\hspace{8pt}}r}
        \toprule
        \textbf{Method} & \textbf{AE2.0 LC} & \textbf{ArenaHard} \\
        \midrule
        Alpaca-5k (baseline) & 19.50 & 22.1 (-2.1, 1.9) \\
        \midrule
        Evol-Instruct & 36.34 & 38.7 (-2.8, 2.1) \\
        Codeclm  & 30.64 & 37.1 (-2.2, 2.4) \\
        Auto-Instruct-Evol & 31.65 & 35.6 (-2.0, 2.0) \\
        Tree-Instruct & 32.71 & 33.6 (-2.3, 1.9) \\
        \rowcolor{blue!10} \mname & \textbf{42.76} & \textbf{41.2 (-2.3, 2.1)} \\
        \bottomrule
    \end{tabular}}
    \caption{Performance comparison with different instruction methods using Qwen2.5-72B-instruct as teacher model. AE2.0 LC represents AlpacaEval2.0 length-controlled win rate.}
    \label{tab:teacher_model}
\end{table}

\section{Conclusion}
In this paper, we propose \mname, a novel framework for enhancing instruction complexity through structured semantic compression and controlled difficulty augmentation. By operating in a compressed tag space rather than raw text, our approach enables more precise and controllable instruction enhancement. Empirical results on AlpacaEval 2 demonstrate that \mname outperforms existing instruction augmentation methods. Our analysis shows that structured semantic manipulation enables controlled progression of instruction difficulty while maintaining coherence across different synthesis frameworks. The effectiveness of our approach suggests that structured semantic manipulation is a promising direction for instruction optimization in language models.

\clearpage
\newpage

\section*{Limitations}
Despite the effectiveness of our approach, we identify several limitations. First, our semantic compression process may lose some nuanced information during the discretization of continuous semantic spaces into discrete tags. Second, using response length as a proxy for instruction quality in tag utility estimation might not fully capture all aspects of instruction complexity. Third, our framework may introduce some noise during the tag expansion process, as we do not incorporate explicit filtering mechanisms to validate the semantic coherence of newly generated tags. Finally, the generalizability of our approach may be limited as our current evaluation focuses primarily on general-domain instructions; future work should explore the effectiveness of \mname in specialized domains such as code generation and mathematical reasoning tasks.

\section*{Acknowledgements}
This project was supported by National Natural Science Foundation of China (No. 62306132) and Guangdong Basic and Applied Basic Research Foundation (No. 2025A1515011564). 
The paper was also supported by the Key Project of the Shanghai Municipal Education Commission's AI-Enabled Research Paradigm Reform and Discipline Leap Program (Development of a Domain-Specific Large Language Model in the Field of Urban and Rural Planning for Enhancing Spatial Cognition and Decision-Making Capabilities) and by the Fundamental Research Funds for the Central Universities (22120250239).
This work was done by He Zhu during his internship at SUSTech. We thank the anonymous reviewers for their insightful feedbacks on this work.

\bibliography{custom}

\begin{thebibliography}{55}
\providecommand{\natexlab}[1]{#1}

\bibitem[{Abdin et~al.(2024)Abdin, Aneja, Behl, Bubeck, Eldan, Gunasekar et~al.}]{phi4}
Marah Abdin, Jyoti Aneja, Harkirat Behl, Sébastien Bubeck, Ronen Eldan, Suriya Gunasekar, et~al. 2024.
\newblock \href {https://arxiv.org/abs/2412.08905} {Phi-4 technical report}.
\newblock \emph{Preprint}, arXiv:2412.08905.

\bibitem[{An et~al.(2024)An, Chen, Sun, and Li}]{sentencevae}
Hongjun An, Yifan Chen, Zhe Sun, and Xuelong Li. 2024.
\newblock \href {https://arxiv.org/abs/2408.00655} {Sentencevae: Enable next-sentence prediction for large language models with faster speed, higher accuracy and longer context}.
\newblock \emph{Preprint}, arXiv:2408.00655.

\bibitem[{Cao et~al.(2023)Cao, Kang, Wang, and Sun}]{instruct-mining}
Yihan Cao, Yanbin Kang, Chi Wang, and Lichao Sun. 2023.
\newblock Instruction mining: Instruction data selection for tuning large language models.
\newblock \emph{arXiv preprint arXiv:2307.06290}.

\bibitem[{Chung et~al.(2022)Chung, Hou, Longpre, Zoph, Tay, Fedus, Li et~al.}]{flanv2}
Hyung~Won Chung, Le~Hou, Shayne Longpre, Barret Zoph, Yi~Tay, William Fedus, Yunxuan Li, et~al. 2022.
\newblock \href {https://arxiv.org/abs/2210.11416} {Scaling instruction-finetuned language models}.
\newblock \emph{Preprint}, arXiv:2210.11416.

\bibitem[{DeepSeek-AI et~al.(2024)DeepSeek-AI, Liu, Feng, Xue, Wang, Wu, Lu, and et.al.}]{deepseekv3}
DeepSeek-AI, Aixin Liu, Bei Feng, Bing Xue, Bingxuan Wang, Bochao Wu, Chengda Lu, and et.al. 2024.
\newblock \href {https://arxiv.org/abs/2412.19437} {Deepseek-v3 technical report}.
\newblock \emph{Preprint}, arXiv:2412.19437.

\bibitem[{Didolkar et~al.(2024)Didolkar, Goyal, Ke, Guo, Valko, Lillicrap, Rezende, Bengio, Mozer, and Arora}]{metacognitive}
Aniket Didolkar, Anirudh Goyal, Nan~Rosemary Ke, Siyuan Guo, Michal Valko, Timothy Lillicrap, Danilo Rezende, Yoshua Bengio, Michael Mozer, and Sanjeev Arora. 2024.
\newblock \href {https://arxiv.org/abs/2405.12205} {Metacognitive capabilities of llms: An exploration in mathematical problem solving}.
\newblock \emph{Preprint}, arXiv:2405.12205.

\bibitem[{Du et~al.(2023)Du, Zong, and Zhang}]{du2023mods}
Qianlong Du, Chengqing Zong, and Jiajun Zhang. 2023.
\newblock \href {https://arxiv.org/abs/2311.15653} {Mods: Model-oriented data selection for instruction tuning}.
\newblock \emph{Preprint}, arXiv:2311.15653.

\bibitem[{Dubois et~al.(2024)Dubois, Galambosi, Liang, and Hashimoto}]{lengthcontrolledalpaca}
Yann Dubois, Balázs Galambosi, Percy Liang, and Tatsunori~B. Hashimoto. 2024.
\newblock \href {https://arxiv.org/abs/2404.04475} {Length-controlled alpacaeval: A simple way to debias automatic evaluators}.
\newblock \emph{Preprint}, arXiv:2404.04475.

\bibitem[{Ge et~al.(2024)Ge, Chan, Wang, Yu, Mi, and Yu}]{1mperson}
Tao Ge, Xin Chan, Xiaoyang Wang, Dian Yu, Haitao Mi, and Dong Yu. 2024.
\newblock \href {https://arxiv.org/abs/2406.20094} {Scaling synthetic data creation with 1,000,000,000 personas}.
\newblock \emph{Preprint}, arXiv:2406.20094.

\bibitem[{Goldshmidt and Horovicz(2024)}]{tokenshap}
Roni Goldshmidt and Miriam Horovicz. 2024.
\newblock \href {https://arxiv.org/abs/2407.10114} {Tokenshap: Interpreting large language models with monte carlo shapley value estimation}.
\newblock \emph{Preprint}, arXiv:2407.10114.

\bibitem[{Grattafiori et~al.(2024)Grattafiori, Dubey, Jauhri, Pandey, Kadian, Al-Dahle, and et.al.}]{llama3}
Aaron Grattafiori, Abhimanyu Dubey, Abhinav Jauhri, Abhinav Pandey, Abhishek Kadian, Ahmad Al-Dahle, and Aiesha~Letman et.al. 2024.
\newblock \href {https://arxiv.org/abs/2407.21783} {The llama 3 herd of models}.
\newblock \emph{Preprint}, arXiv:2407.21783.

\bibitem[{Guo et~al.(2025)Guo, Yang, Zhang, Song, Zhang, Xu, Zhu, Ma, Wang, Bi et~al.}]{deepseek-r1}
Daya Guo, Dejian Yang, Haowei Zhang, Junxiao Song, Ruoyu Zhang, Runxin Xu, Qihao Zhu, Shirong Ma, Peiyi Wang, Xiao Bi, et~al. 2025.
\newblock Deepseek-r1: Incentivizing reasoning capability in llms via reinforcement learning.
\newblock \emph{arXiv preprint arXiv:2501.12948}.

\bibitem[{Hao et~al.(2024)Hao, Sukhbaatar, Su, Li, Hu, Weston, and Tian}]{coconut}
Shibo Hao, Sainbayar Sukhbaatar, DiJia Su, Xian Li, Zhiting Hu, Jason Weston, and Yuandong Tian. 2024.
\newblock \href {https://arxiv.org/abs/2412.06769} {Training large language models to reason in a continuous latent space}.
\newblock \emph{Preprint}, arXiv:2412.06769.

\bibitem[{Jiang et~al.(2023)Jiang, Sablayrolles, Mensch, Bamford, Chaplot, and et.al.}]{jiang2023mistral7b}
Albert~Q. Jiang, Alexandre Sablayrolles, Arthur Mensch, Chris Bamford, Devendra~Singh Chaplot, and et.al. 2023.
\newblock \href {https://arxiv.org/abs/2310.06825} {Mistral 7b}.
\newblock \emph{Preprint}, arXiv:2310.06825.

\bibitem[{Kaur et~al.(2024)Kaur, Park, Goyal, and Arora}]{instructskillmix}
Simran Kaur, Simon Park, Anirudh Goyal, and Sanjeev Arora. 2024.
\newblock \href {https://arxiv.org/abs/2408.14774} {Instruct-skillmix: A powerful pipeline for llm instruction tuning}.
\newblock \emph{Preprint}, arXiv:2408.14774.

\bibitem[{Kaur et~al.(2025)Kaur, Park, Goyal, and Arora}]{kaur2025instructskillmix}
Simran Kaur, Simon Park, Anirudh Goyal, and Sanjeev Arora. 2025.
\newblock \href {https://arxiv.org/abs/2408.14774} {Instruct-skillmix: A powerful pipeline for llm instruction tuning}.
\newblock \emph{Preprint}, arXiv:2408.14774.

\bibitem[{Kemp et~al.(2018)Kemp, Xu, and Regier}]{kemp2018semantic}
Charles Kemp, Yang Xu, and Terry Regier. 2018.
\newblock \href {https://doi.org/10.1146/annurev-linguistics-011817-045406} {Semantic typology and efficient communication}.
\newblock \emph{Annual Review of Linguistics}, 4:109--128.

\bibitem[{Kingma and Welling(2013)}]{kingma2013auto}
Diederik~P Kingma and Max Welling. 2013.
\newblock Auto-encoding variational bayes.
\newblock \emph{arXiv preprint arXiv:1312.6114}.

\bibitem[{Kingma and Welling(2022)}]{vae}
Diederik~P Kingma and Max Welling. 2022.
\newblock \href {https://arxiv.org/abs/1312.6114} {Auto-encoding variational bayes}.
\newblock \emph{Preprint}, arXiv:1312.6114.

\bibitem[{Kuhn et~al.(2023)Kuhn, Gal, and Farquhar}]{semanticentropy}
Lorenz Kuhn, Yarin Gal, and Sebastian Farquhar. 2023.
\newblock \href {https://arxiv.org/abs/2302.09664} {Semantic uncertainty: Linguistic invariances for uncertainty estimation in natural language generation}.
\newblock \emph{Preprint}, arXiv:2302.09664.

\bibitem[{Lambert et~al.(2025)Lambert, Morrison, Pyatkin, Huang, Ivison, Brahman et~al.}]{tulu3}
Nathan Lambert, Jacob Morrison, Valentina Pyatkin, Shengyi Huang, Hamish Ivison, Faeze Brahman, et~al. 2025.
\newblock \href {https://arxiv.org/abs/2411.15124} {Tulu 3: Pushing frontiers in open language model post-training}.
\newblock \emph{Preprint}, arXiv:2411.15124.

\bibitem[{Li et~al.(2024{\natexlab{a}})Li, Zhang, Li, Chen, Chen, Cheng, Wang, Zhou, and Xiao}]{cherryllm}
Ming Li, Yong Zhang, Zhitao Li, Jiuhai Chen, Lichang Chen, Ning Cheng, Jianzong Wang, Tianyi Zhou, and Jing Xiao. 2024{\natexlab{a}}.
\newblock \href {https://arxiv.org/abs/2308.12032} {From quantity to quality: Boosting llm performance with self-guided data selection for instruction tuning}.
\newblock \emph{Preprint}, arXiv:2308.12032.

\bibitem[{Li et~al.(2024{\natexlab{b}})Li, Chiang, Frick, Dunlap, Wu, Zhu, Gonzalez, and Stoica}]{arenahard}
Tianle Li, Wei-Lin Chiang, Evan Frick, Lisa Dunlap, Tianhao Wu, Banghua Zhu, Joseph~E. Gonzalez, and Ion Stoica. 2024{\natexlab{b}}.
\newblock \href {https://arxiv.org/abs/2406.11939} {From crowdsourced data to high-quality benchmarks: Arena-hard and benchbuilder pipeline}.
\newblock \emph{Preprint}, arXiv:2406.11939.

\bibitem[{Li et~al.(2023)Li, Zhang, Dubois, Taori, Gulrajani, Guestrin, Liang, and Hashimoto}]{alpaca_eval}
Xuechen Li, Tianyi Zhang, Yann Dubois, Rohan Taori, Ishaan Gulrajani, Carlos Guestrin, Percy Liang, and Tatsunori~B. Hashimoto. 2023.
\newblock Alpacaeval: An automatic evaluator of instruction-following models.
\newblock \url{https://github.com/tatsu-lab/alpaca_eval}.

\bibitem[{Liu et~al.(2023{\natexlab{a}})Liu, Zeng, He, Jiang, and He}]{liu2023what}
Wei Liu, Weihao Zeng, Keqing He, Yong Jiang, and Junxian He. 2023{\natexlab{a}}.
\newblock \href {https://arxiv.org/abs/2312.15685} {What makes good data for alignment? a comprehensive study of automatic data selection in instruction tuning}.
\newblock \emph{Preprint}, arXiv:2312.15685.

\bibitem[{Liu et~al.(2023{\natexlab{b}})Liu, Zeng, He, Jiang, and He}]{liu2023deita}
Wei Liu, Weihao Zeng, Keqing He, Yong Jiang, and Junxian He. 2023{\natexlab{b}}.
\newblock \href {https://arxiv.org/abs/2312.15685} {What makes good data for alignment? a comprehensive study of automatic data selection in instruction tuning}.
\newblock \emph{Preprint}, arXiv:2312.15685.

\bibitem[{Lu et~al.(2023)Lu, Yuan, Yuan, Lin, Lin, Tan, and Zhou}]{instag}
Keming Lu, Hongyi Yuan, Zheng Yuan, Runji Lin, Junyang Lin, Chuanqi Tan, and Chang Zhou. 2023.
\newblock \# instag: Instruction tagging for diversity and complexity analysis.
\newblock \emph{arXiv preprint arXiv:2308.07074}.

\bibitem[{Nvidia et~al.(2024)Nvidia, :, Adler, Agarwal, Aithal, Anh, Bhattacharya, Brundyn, Casper, Catanzaro, Clay, Cohen, Das, Dattagupta, Delalleau, Derczynski, Dong, Egert, Evans, Ficek, Fridman, Ghosh, Ginsburg, Gitman, Grzegorzek, Hero, Huang, Jawa, Jennings, Jhunjhunwala, Kamalu, Khan, Kuchaiev, LeGresley, Li, Liu, Liu, Long, Mahabaleshwarkar, Majumdar, Maki, Martinez, de~Melo, Moshkov, Narayanan, Narenthiran, Navarro, Nguyen, Nitski, Noroozi, Nutheti, Parisien, Parmar, Patwary, Pawelec, Ping, Prabhumoye, Roy, Saar, Sabavat, Satheesh, Scowcroft, Sewall, Shamis, Shen, Shoeybi, Sizer, Smelyanskiy, Soares, Sreedhar, Su, Subramanian, Sun, Toshniwal, Wang, Wang, You, Zeng, Zhang, Zhang, Zhang, Zhang, and Zhu}]{nemotron4340btechnicalreport}
Nvidia, :, Bo~Adler, Niket Agarwal, Ashwath Aithal, Dong~H. Anh, Pallab Bhattacharya, Annika Brundyn, Jared Casper, Bryan Catanzaro, Sharon Clay, Jonathan Cohen, Sirshak Das, Ayush Dattagupta, Olivier Delalleau, Leon Derczynski, Yi~Dong, Daniel Egert, Ellie Evans, Aleksander Ficek, Denys Fridman, Shaona Ghosh, Boris Ginsburg, Igor Gitman, Tomasz Grzegorzek, Robert Hero, Jining Huang, Vibhu Jawa, Joseph Jennings, Aastha Jhunjhunwala, John Kamalu, Sadaf Khan, Oleksii Kuchaiev, Patrick LeGresley, Hui Li, Jiwei Liu, Zihan Liu, Eileen Long, Ameya~Sunil Mahabaleshwarkar, Somshubra Majumdar, James Maki, Miguel Martinez, Maer~Rodrigues de~Melo, Ivan Moshkov, Deepak Narayanan, Sean Narenthiran, Jesus Navarro, Phong Nguyen, Osvald Nitski, Vahid Noroozi, Guruprasad Nutheti, Christopher Parisien, Jupinder Parmar, Mostofa Patwary, Krzysztof Pawelec, Wei Ping, Shrimai Prabhumoye, Rajarshi Roy, Trisha Saar, Vasanth Rao~Naik Sabavat, Sanjeev Satheesh, Jane~Polak Scowcroft, Jason Sewall, Pavel Shamis, Gerald Shen, Mohammad
  Shoeybi, Dave Sizer, Misha Smelyanskiy, Felipe Soares, Makesh~Narsimhan Sreedhar, Dan Su, Sandeep Subramanian, Shengyang Sun, Shubham Toshniwal, Hao Wang, Zhilin Wang, Jiaxuan You, Jiaqi Zeng, Jimmy Zhang, Jing Zhang, Vivienne Zhang, Yian Zhang, and Chen Zhu. 2024.
\newblock \href {https://arxiv.org/abs/2406.11704} {Nemotron-4 340b technical report}.
\newblock \emph{Preprint}, arXiv:2406.11704.

\bibitem[{OpenAI et~al.(2024)OpenAI, Achiam, Adler, Agarwal, Ahmad, Akkaya, Aleman, and et.al.}]{gpt4}
OpenAI, Josh Achiam, Steven Adler, Sandhini Agarwal, Lama Ahmad, Ilge Akkaya, Florencia~Leoni Aleman, and Diogo~Almeida et.al. 2024.
\newblock \href {https://arxiv.org/abs/2303.08774} {Gpt-4 technical report}.
\newblock \emph{Preprint}, arXiv:2303.08774.

\bibitem[{Qwen et~al.(2024)Qwen, :, Yang, Yang, Zhang, Hui, Zheng, Yu et~al.}]{qwen25}
Qwen, :, An~Yang, Baosong Yang, Beichen Zhang, Binyuan Hui, Bo~Zheng, Bowen Yu, et~al. 2024.
\newblock \href {https://arxiv.org/abs/2412.15115} {Qwen2.5 technical report}.
\newblock \emph{Preprint}, arXiv:2412.15115.

\bibitem[{Rafailov et~al.(2024)Rafailov, Sharma, Mitchell, Ermon, Manning, and Finn}]{dpo}
Rafael Rafailov, Archit Sharma, Eric Mitchell, Stefano Ermon, Christopher~D. Manning, and Chelsea Finn. 2024.
\newblock \href {https://arxiv.org/abs/2305.18290} {Direct preference optimization: Your language model is secretly a reward model}.
\newblock \emph{Preprint}, arXiv:2305.18290.

\bibitem[{Reimers and Gurevych(2019)}]{reimers2019sentence}
Nils Reimers and Iryna Gurevych. 2019.
\newblock \href {https://arxiv.org/abs/1908.10084} {Sentence-bert: Sentence embeddings using siamese bert-networks}.
\newblock \emph{Preprint}, arXiv:1908.10084.

\bibitem[{Shapley(1953)}]{shapley1953value}
Lloyd~S Shapley. 1953.
\newblock A value for n-person games.
\newblock \emph{Contributions to the Theory of Games}, 2(28):307--317.

\bibitem[{Shen(2024)}]{shen2024rethinkingdataselectionsupervised}
Ming Shen. 2024.
\newblock \href {https://arxiv.org/abs/2402.06094} {Rethinking data selection for supervised fine-tuning}.
\newblock \emph{Preprint}, arXiv:2402.06094.

\bibitem[{Taori et~al.(2023)Taori, Gulrajani, Zhang, Dubois, Li, Guestrin, Liang, and Hashimoto}]{alpaca}
Rohan Taori, Ishaan Gulrajani, Tianyi Zhang, Yann Dubois, Xuechen Li, Carlos Guestrin, Percy Liang, and Tatsunori~B. Hashimoto. 2023.
\newblock Stanford alpaca: An instruction-following llama model.
\newblock \url{https://github.com/tatsu-lab/stanford_alpaca}.

\bibitem[{team et~al.(2024)team, Barrault, Duquenne, Elbayad, Kozhevnikov, Alastruey, Andrews, Coria, Couairon, Costa-jussà, Dale, Elsahar, Heffernan, Janeiro, Tran, Ropers, Sánchez, Roman, Mourachko, Saleem, and Schwenk}]{largeconceptmodel}
LCM team, Loïc Barrault, Paul-Ambroise Duquenne, Maha Elbayad, Artyom Kozhevnikov, Belen Alastruey, Pierre Andrews, Mariano Coria, Guillaume Couairon, Marta~R. Costa-jussà, David Dale, Hady Elsahar, Kevin Heffernan, João~Maria Janeiro, Tuan Tran, Christophe Ropers, Eduardo Sánchez, Robin~San Roman, Alexandre Mourachko, Safiyyah Saleem, and Holger Schwenk. 2024.
\newblock \href {https://arxiv.org/abs/2412.08821} {Large concept models: Language modeling in a sentence representation space}.
\newblock \emph{Preprint}, arXiv:2412.08821.

\bibitem[{Wang et~al.(2024{\natexlab{a}})Wang, Cassano, Wu, Bai, Song, Nath, Han, Hendryx, Yue, and Zhang}]{plansearch}
Evan Wang, Federico Cassano, Catherine Wu, Yunfeng Bai, Will Song, Vaskar Nath, Ziwen Han, Sean Hendryx, Summer Yue, and Hugh Zhang. 2024{\natexlab{a}}.
\newblock \href {https://arxiv.org/abs/2409.03733} {Planning in natural language improves llm search for code generation}.
\newblock \emph{Preprint}, arXiv:2409.03733.

\bibitem[{Wang et~al.(2023)Wang, Kordi, Mishra, Liu, Smith, Khashabi, and Hajishirzi}]{self_instruct}
Yizhong Wang, Yeganeh Kordi, Swaroop Mishra, Alisa Liu, Noah~A. Smith, Daniel Khashabi, and Hannaneh Hajishirzi. 2023.
\newblock \href {https://arxiv.org/abs/2212.10560} {Self-instruct: Aligning language models with self-generated instructions}.
\newblock \emph{Preprint}, arXiv:2212.10560.

\bibitem[{Wang et~al.(2024{\natexlab{b}})Wang, Li, Perot, Le, Miao, Zhang, Lee, and Pfister}]{codeclm}
Zifeng Wang, Chun-Liang Li, Vincent Perot, Long~T. Le, Jin Miao, Zizhao Zhang, Chen-Yu Lee, and Tomas Pfister. 2024{\natexlab{b}}.
\newblock \href {https://arxiv.org/abs/2404.05875} {Codeclm: Aligning language models with tailored synthetic data}.
\newblock \emph{Preprint}, arXiv:2404.05875.

\bibitem[{Wettig et~al.(2024)Wettig, Gupta, Malik, and Chen}]{wettig2024qurating}
Alexander Wettig, Aatmik Gupta, Saumya Malik, and Danqi Chen. 2024.
\newblock \href {https://arxiv.org/abs/2402.09739} {Qurating: Selecting high-quality data for training language models}.
\newblock \emph{Preprint}, arXiv:2402.09739.

\bibitem[{Xia et~al.(2024)Xia, Malladi, Gururangan, Arora, and Chen}]{LESS}
Mengzhou Xia, Sadhika Malladi, Suchin Gururangan, Sanjeev Arora, and Danqi Chen. 2024.
\newblock \href {https://arxiv.org/abs/2402.04333} {Less: Selecting influential data for targeted instruction tuning}.
\newblock \emph{Preprint}, arXiv:2402.04333.

\bibitem[{Xu et~al.(2023)Xu, Sun, Zheng, Geng, Zhao, Feng, Tao, and Jiang}]{wizardlm}
Can Xu, Qingfeng Sun, Kai Zheng, Xiubo Geng, Pu~Zhao, Jiazhan Feng, Chongyang Tao, and Daxin Jiang. 2023.
\newblock \href {https://arxiv.org/abs/2304.12244} {Wizardlm: Empowering large language models to follow complex instructions}.
\newblock \emph{Preprint}, arXiv:2304.12244.

\bibitem[{Xu et~al.(2024)Xu, Jiang, Niu, Deng, Poovendran, Choi, and Lin}]{magpie}
Zhangchen Xu, Fengqing Jiang, Luyao Niu, Yuntian Deng, Radha Poovendran, Yejin Choi, and Bill~Yuchen Lin. 2024.
\newblock \href {https://arxiv.org/abs/2406.08464} {Magpie: Alignment data synthesis from scratch by prompting aligned llms with nothing}.
\newblock \emph{Preprint}, arXiv:2406.08464.

\bibitem[{Yang et~al.(2024{\natexlab{a}})Yang, Yang, Hui, Zheng, Yu, Zhou, Li, Li, Liu, Huang et~al.}]{qwen2}
An~Yang, Baosong Yang, Binyuan Hui, Bo~Zheng, Bowen Yu, Chang Zhou, Chengpeng Li, Chengyuan Li, Dayiheng Liu, Fei Huang, et~al. 2024{\natexlab{a}}.
\newblock Qwen2 technical report.
\newblock \emph{arXiv preprint arXiv:2407.10671}.

\bibitem[{Yang et~al.(2024{\natexlab{b}})Yang, Pang, Feng, Wang, Chen, Zhu, and Liu}]{self-distillation}
Zhaorui Yang, Tianyu Pang, Haozhe Feng, Han Wang, Wei Chen, Minfeng Zhu, and Qian Liu. 2024{\natexlab{b}}.
\newblock \href {https://arxiv.org/abs/2402.13669} {Self-distillation bridges distribution gap in language model fine-tuning}.
\newblock \emph{Preprint}, arXiv:2402.13669.

\bibitem[{Ye et~al.(2025)Ye, Huang, Xiao, Chern, Xia, and Liu}]{limo}
Yixin Ye, Zhen Huang, Yang Xiao, Ethan Chern, Shijie Xia, and Pengfei Liu. 2025.
\newblock \href {https://arxiv.org/abs/2502.03387} {Limo: Less is more for reasoning}.
\newblock \emph{Preprint}, arXiv:2502.03387.

\bibitem[{Yu et~al.(2024)Yu, Xu, Weston, and Kulikov}]{distilled}
Ping Yu, Jing Xu, Jason Weston, and Ilia Kulikov. 2024.
\newblock \href {https://arxiv.org/abs/2407.06023} {Distilling system 2 into system 1}.
\newblock \emph{Preprint}, arXiv:2407.06023.

\bibitem[{Zeng et~al.(2024)Zeng, Xu, Zhao, Lou, and Chen}]{auto-instruct-evol}
Weihao Zeng, Can Xu, Yingxiu Zhao, Jian-Guang Lou, and Weizhu Chen. 2024.
\newblock \href {https://arxiv.org/abs/2406.00770} {Automatic instruction evolving for large language models}.
\newblock \emph{Preprint}, arXiv:2406.00770.

\bibitem[{Zhao et~al.(2024)Zhao, Andriushchenko, Croce, and Flammarion}]{zhao2024long}
Hao Zhao, Maksym Andriushchenko, Francesco Croce, and Nicolas Flammarion. 2024.
\newblock \href {https://arxiv.org/abs/2402.04833} {Long is more for alignment: A simple but tough-to-beat baseline for instruction fine-tuning}.
\newblock \emph{Preprint}, arXiv:2402.04833.

\bibitem[{Zhao et~al.(2023)Zhao, Yu, Hui, Yu, Huang, Li, and Zhang}]{tree-instruct}
Yingxiu Zhao, Bowen Yu, Binyuan Hui, Haiyang Yu, Fei Huang, Yongbin Li, and Nevin~L Zhang. 2023.
\newblock A preliminary study of the intrinsic relationship between complexity and alignment.
\newblock \emph{arXiv preprint arXiv:2308.05696}.

\bibitem[{Zheng et~al.(2024{\natexlab{a}})Zheng, Mishra, Chen, Cheng, Chi, Le, and Zhou}]{stepback}
Huaixiu~Steven Zheng, Swaroop Mishra, Xinyun Chen, Heng-Tze Cheng, Ed~H. Chi, Quoc~V Le, and Denny Zhou. 2024{\natexlab{a}}.
\newblock \href {https://arxiv.org/abs/2310.06117} {Take a step back: Evoking reasoning via abstraction in large language models}.
\newblock \emph{Preprint}, arXiv:2310.06117.

\bibitem[{Zheng et~al.(2024{\natexlab{b}})Zheng, Chiang, Sheng, Zhuang, Wu, Zhuang, Lin, Li, Li, Xing et~al.}]{zheng2024judging}
Lianmin Zheng, Wei-Lin Chiang, Ying Sheng, Siyuan Zhuang, Zhanghao Wu, Yonghao Zhuang, Zi~Lin, Zhuohan Li, Dacheng Li, Eric Xing, et~al. 2024{\natexlab{b}}.
\newblock Judging llm-as-a-judge with mt-bench and chatbot arena.
\newblock \emph{Advances in Neural Information Processing Systems}, 36.

\bibitem[{Zheng et~al.(2023)Zheng, Chiang, Sheng, Zhuang, Wu, Zhuang, Lin, Li, Li, Xing, Zhang, Gonzalez, and Stoica}]{zheng2023judgingllmasajudgemtbenchchatbot}
Lianmin Zheng, Wei-Lin Chiang, Ying Sheng, Siyuan Zhuang, Zhanghao Wu, Yonghao Zhuang, Zi~Lin, Zhuohan Li, Dacheng Li, Eric~P. Xing, Hao Zhang, Joseph~E. Gonzalez, and Ion Stoica. 2023.
\newblock \href {https://arxiv.org/abs/2306.05685} {Judging llm-as-a-judge with mt-bench and chatbot arena}.
\newblock \emph{Preprint}, arXiv:2306.05685.

\bibitem[{Zhou et~al.(2023)Zhou, Liu, Xu, Iyer, Sun, Mao, Ma, Efrat, Yu, Yu, Zhang, Ghosh, Lewis, Zettlemoyer, and Levy}]{lima}
Chunting Zhou, Pengfei Liu, Puxin Xu, Srini Iyer, Jiao Sun, Yuning Mao, Xuezhe Ma, Avia Efrat, Ping Yu, Lili Yu, Susan Zhang, Gargi Ghosh, Mike Lewis, Luke Zettlemoyer, and Omer Levy. 2023.
\newblock \href {https://arxiv.org/abs/2305.11206} {Lima: Less is more for alignment}.
\newblock \emph{Preprint}, arXiv:2305.11206.

\bibitem[{Zhu et~al.(2024)Zhu, Su, Lun, Tao, Zhang, Fan, and Chen}]{fanno}
He~Zhu, Junyou Su, Tianle Lun, Yicheng Tao, Wenjia Zhang, Zipei Fan, and Guanhua Chen. 2024.
\newblock \href {https://arxiv.org/abs/2408.01323} {Fanno: Augmenting high-quality instruction data with open-sourced llms only}.
\newblock \emph{Preprint}, arXiv:2408.01323.

\end{thebibliography}

\appendix
\onecolumn

\section{\mname Prompt Template Demonstration} \label{appendix:prompt_demo}

\subsection{Encode/Decode Prompts for I2T and T2I} \label{appendix:encode_decode}
This section presents the prompt templates used for instruction-to-tag (I2T) encoding and tag-to-instruction (T2I) decoding. The I2T prompt extracts semantic tags from instructions, while the T2I prompt generates enhanced instructions based on tags and original questions.

\begin{figure*}[!ht]
\centering
\begin{tcolorbox}[title=Semantic Tag Extraction Prompt Template for I2T Encoding]
\small
You are a semantic analysis expert. Your task is to extract exactly three essential tags that capture the core semantic concepts of the given instruction or question. The tags should be:\\
- Concise (1–2 words each)\\
- Hierarchically ordered by importance\\
- Generalizable across similar tasks\\

\#\#\# Guidelines:\\
1. Focus on action-oriented concepts\\
2. Avoid redundant or overlapping tags\\
3. Use standard terminology when possible\\

\#\#\# Examples:\\
\textnormal{[Input]}\\
Describe a situation where team collaboration improved the outcome of a project.\\
\textnormal{[Tags]}\\
teamwork\_experience, project\_outcomes, success\_factors\\

\textnormal{[Input]}\\
What strategies can be used to improve time management in a busy work environment?\\
\textnormal{[Tags]}\\
productivity\_methods, workload\_optimization, efficiency\_tactics\\

\textnormal{[Input]}\\
Outline the steps required to create a successful marketing campaign for a new product launch.\\
\textnormal{[Tags]}\\
campaign\_planning, market\_strategy, launch\_execution\\

\textnormal{[Input]}\\
In the context of climate change adaptation, analyze how urban planning strategies can be modified to create resilient cities that can withstand extreme weather events while maintaining economic growth and social equity for their residents over the next 50 years.\\
\textnormal{[Tags]}\\
urban resilience, climate adaptation, sustainable development\\

\textnormal{[Input]}\\
Design a comprehensive employee training program for a multinational corporation that addresses cultural sensitivity, remote work effectiveness, and digital tool proficiency while ensuring consistent skill development across different time zones and accounting for various learning styles and language barriers.\\
\textnormal{[Tags]}\\
corporate training, global workforce, skill development\\

\textnormal{[Input]}\\
Develop a detailed analysis of how artificial intelligence implementations in healthcare systems can improve patient outcomes while considering privacy concerns, medical ethics, and the integration challenges with existing hospital infrastructure and staff training requirements.\\
\textnormal{[Tags]}\\
healthcare AI, medical ethics, system integration\\

\#\#\# Task:\\
Given the following instruction, provide exactly three semantic tags following the above format and guidelines:\\

\textnormal{[Input]}\\
\{instruction\}\\

\textnormal{[Output Format]}\\
tag1, tag2, tag3\\

\textnormal{[Tags]}
\end{tcolorbox}
\caption{Prompt template for extracting semantic tags from instructions, with examples demonstrating tag extraction across diverse domains}
\label{fig:encode_prompt}
\end{figure*}

\begin{figure*}[htbp]
    \centering
    \begin{tcolorbox}[title=Tag-to-Instruction Prompt Template for T2I Decoding]
    \small
    Using the provided tags and original question, generate a new version of the question that is more complex and thought-provoking.\\
    
    \#\#\# Requirements:\\
    1. \textbf{Tag Selection}: Choose a subset of the tags that best enhance the depth of the question.\\
    2. \textbf{New Question Complexity}: The new question should be more challenging than the original, requiring deeper thought, but it should be solvable. Avoid simply adding length; instead, focus on making the question more insightful and intellectually engaging.\\
    3. \textbf{Solvability}: Ensure the new question remains clear and achievable.\\
    
    \#\#\# Output Format:\\
    \textnormal{[Tags]}\\
    Here are the existing tags.\\
    
    \textnormal{[Original Question]}\\
    Here is the original question.\\
    
    Output:\\
    \textnormal{[New Question]}\\
    Provide the new, more complex question.\\
    
    \#\#\# Your Task:\\
    \textnormal{[Tags]}\\
    \{tags\}\\
    
    \textnormal{[Original Question]}\\
    \{question\}\\
    
    Output:\\
    \textnormal{[New Question]}
    \end{tcolorbox}
    \caption{Template for generating enhanced instructions from semantic tags while maintaining clarity and solvability}
    \label{fig:decode_prompt}
\end{figure*}

\begin{figure*}[htbp]
    \centering
    \begin{tcolorbox}[title=Tag Expansion and Complexity Enhancement Prompt Template]
    \small
    Please generate only one new tag based on the existing tags and the given task or question. Then, using all the tags, create a new, more challenging version of the task or question.\\
    
    \#\#\# Important:\\
    The new tag should differ from the previous tags and relate to the context of the question.\\
    
    \#\#\# Format:\\
    \texttt{[Tags]}: Here are the existing tags.\\
    \texttt{[Original Question]}: Here is the original question.\\
    Output:\\
    \texttt{[New Tag]}: Here is the new tag.\\
    \texttt{[New Question]}: Here is the new question.\\
    
    \#\#\# Your Task:\\
    \texttt{[Tags]}: \{tags\}\\
    \texttt{[Original Question]}: \{question\}\\
    Output:
    \end{tcolorbox}
    \caption{Template for expanding tag set and enhancing question complexity through contextual tag addition}
    \label{appendix:adding_tag_template}
\end{figure*}

\begin{figure*}[htbp]
    \centering
    \begin{tcolorbox}[title=Response Generation Prompt Template]
    \small
    Below is an instruction that describes a task, write a response that appropriately completes the request.\\
    
    \#\#\# Instruction:\\
    \{instruction\}\\
    
    \#\#\# Response:
    \end{tcolorbox}
    \caption{Simple template for generating responses to task instructions}
    \label{fig:response_prompt}
\end{figure*}

\section{Experimental Setup Details}
\label{appendix:exp_setup}

\subsection{Dataset Specifications}
We utilized Alpaca-Clean~\citep{alpaca}, which contains high-quality instruction-response pairs generated by GPT-4 and manually filtered by human annotators. From this dataset, we randomly sampled 5K instructions to create Alpaca-5k as our initial instruction set.

\subsection{Model Architecture and Training Protocol}
Our implementation uses the following specifications:

\paragraph{Base Models}
We employed LLaMA-3-8B and LLaMA-3.2-3B as base models, with Ministral-Instruct-8b \citep{jiang2023mistral7b} serving as the teacher model.

\paragraph{Training Configuration}
The training protocol followed these parameters: Training Duration of 3 epochs, Learning Rate starting at $2 \times 10^{-5}$ with cosine schedule, utilizing 8 GPUs with 80GB memory each, global Batch Size of 128, and maximum Sequence Length of 2048 tokens.

\paragraph{Instruction Processing}
For instruction processing, we used the Alpaca template for single-turn settings, performed tag extraction following methodology from \citet{instag}, conducted tag reward estimation using instructions from \citet{magpie}, and implemented policy model training for the tag expansion task. For additional implementation details, refer to Appendix~\ref{appendix:rl_expansion}.

\section{Detailed Description of Baselines}
\label{appendix:baselines}

\subsection{Methodological Baselines}
We compared our approach against the following methodological baselines:

\begin{itemize}
\item \textbf{Self-Instruct}~\cite{self_instruct}:
This method generates synthetic instruction-following examples automatically to enhance model alignment with human instructions.

\item \textbf{Evol-Instruct}~\cite{wizardlm}:
An iterative instruction evolution framework that progressively increases instruction complexity while maintaining quality, enabling large-scale generation of high-complexity instruction data for LLM training.

\item \textbf{Tree-Instruct}~\cite{tree-instruct}:
This approach systematically enhances instruction complexity by adding nodes to semantic trees, allowing controlled difficulty levels.

\item \textbf{Auto-Instruct-Evol}~\cite{auto-instruct-evol}:
An end-to-end framework that evolves instruction datasets using LLMs without human effort, automatically analyzing and optimizing evolutionary strategies.

\item \textbf{CodecLM}~\cite{codeclm}:
This method employs encode-decode principles with LLMs as Codecs, using metadata to capture target instruction distributions and create tailored instructions.
\end{itemize}

\subsection{Data Baselines}
We also compared against the following datasets:

\begin{itemize}
\item \textbf{Alpaca-Clean}~\cite{alpaca}:
A high-quality instruction-following dataset generated by GPT-4 and manually filtered by human annotators.

\item \textbf{WizardLM-data}~\cite{wizardlm}:
A large-scale instruction-following dataset created through evolved instructions using proprietary LLMs.
\end{itemize}

\section{Tag Expansion Implementation Details}
\label{appendix:rl_expansion}

This section provides implementation details for our tag expansion approach using reinforcement learning to optimize the policy model for high-utility tag generation.

\subsection{Tag Pool Construction}

We construct a comprehensive tag pool using Magpie-500k \cite{magpie} as seed data:

\paragraph{Tag Extraction and Utility Estimation} For each instruction in Magpie-500k, we extract semantic tags using the teacher LLM following Section~\ref{sec:semantic_encoding}. We then compute the utility score for each unique tag using Equation~\ref{eq:reward}:
\[\phi_i = \frac{\sum_{j \in \mathcal{D}_i} |y_j|}{|\mathcal{D}_i|}\]
where $\mathcal{D}_i$ denotes instructions containing tag $i$.

\paragraph{Tag Pool Formation} Based on utility scores, we rank all extracted tags and form the tag pool $\mathcal{T} = \{t_1, t_2, \ldots, t_N\}$ ordered by decreasing utility. We define $\mathcal{T}_{\text{good}}$ as the top 10,000 highest-utility tags and $\mathcal{T}_{\text{bad}}$ as the bottom 10,000 lowest-utility tags.

\subsection{Best-of-N Tag Expansion and Scoring}
\paragraph{Candidate Generation} For each base tag set $z_{\text{base}}$, we follow the prompt template in Figure~\ref{appendix:adding_tag_template} to generate $N=20$ candidate expansion tags using the teacher LLM:
\[z_{\text{new}}^{(i)} \sim p(z|z_{\text{base}}, x), \quad i = 1, 2, \ldots, 20\]

\paragraph{Embedding-based Scoring} We score each candidate tag $z_{\text{new}}^{(i)}$ using embedding similarity with the pre-constructed tag pools. Let $\text{emb}(\cdot)$ denote the embedding function, we compute:
\begin{align}
\text{score}_{\text{good}}^{(i)} &= \frac{1}{|\mathcal{T}_{\text{good}}|} \sum_{t \in \mathcal{T}_{\text{good}}} \cos(\text{emb}(z_{\text{new}}^{(i)}), \text{emb}(t)) \\
\text{score}_{\text{bad}}^{(i)} &= \frac{1}{|\mathcal{T}_{\text{bad}}|} \sum_{t \in \mathcal{T}_{\text{bad}}} \cos(\text{emb}(z_{\text{new}}^{(i)}), \text{emb}(t))
\end{align}

The final score for candidate $i$ is:
\[\text{score}^{(i)} = \text{score}_{\text{good}}^{(i)} - \text{score}_{\text{bad}}^{(i)}\]

The insight is that high-quality tags should be semantically closer to high-utility tags and farther from low-utility tags.

\subsection{Preference Pair Construction}

Based on the computed scores, we construct preference pairs for DPO training. For the chosen tag, we select $z_{\text{chosen}} = \arg\max_{i} \text{score}^{(i)}$ which represents the highest-scoring candidate. For the rejected tag, we randomly sample $z_{\text{rejected}}$ from the remaining candidates $\{z_{\text{new}}^{(j)} : j \neq \arg\max_{i} \text{score}^{(i)}\}$.

\subsection{Policy Model Training}

\paragraph{DPO Training} The policy model is optimized using Direct Preference Optimization on the collected preference pairs:
\[\mathcal{L}_{\text{DPO}} = -\mathbb{E}_{(z_w, z_l)}\left[\log \sigma\left(\beta \log \frac{\pi_\theta(z_w|z_{\text{base}}, x)}{\pi_{\text{ref}}(z_w|z_{\text{base}}, x)} - \beta \log \frac{\pi_\theta(z_l|z_{\text{base}}, x)}{\pi_{\text{ref}}(z_l|z_{\text{base}}, x)}\right)\right]\]

where $z_w$ and $z_l$ denote chosen and rejected tags respectively, $\pi_{\text{ref}}$ is the reference model (teacher LLM), and $\beta$ controls the KL penalty strength. We trained the model using a learning rate of $5 \times 10^{-5}$ with batch size of 128 preference pairs over 1 epochs. The KL penalty coefficient ($\beta$) was set to 0.1, and we used the \texttt{sentence-transformers/all-MiniLM-L6-v2} model for generating embeddings.

While the above process can also be implemented using Proximal Policy Optimization (PPO) with the reward function:
\[R(z_{\text{new}}) = \text{score}_{\text{good}} - \text{score}_{\text{bad}}\]
and the PPO objective:
\[\mathcal{L}_{\text{PPO}} = \mathbb{E}_t\left[\min\left(r_t(\theta)\hat{A}_t, \text{clip}(r_t(\theta), 1-\epsilon, 1+\epsilon)\hat{A}_t\right)\right]\]
where $r_t(\theta) = \frac{\pi_\theta(a_t|s_t)}{\pi_{\text{old}}(a_t|s_t)}$ and $\hat{A}_t$ is the advantage estimate, we choose DPO for its superior efficiency and stability in practice.

\section{Analysis Experiment Setup}
\subsection{Tag Extraction Experiment} \label{appendix:encode_prompt}

We selected 1,000 instructions from alpaca-clean ~\cite{alpaca} and conducted five iterations of tag expansion to obtain a total of 5k instructions. We then followed the same supervised fine-tuning process as in the previous section. Our aim was to evaluate the performance of different encoding strategies in generating tags from instructions. To systematically evaluate different encoding approaches, we designed four template variants that explore different aspects of instruction encoding:

\begin{itemize}
    \item Basic: A minimal template focused on simple tag extraction without additional guidance
    \item Enhanced: A comprehensive template that jointly captures task intent and semantic meaning through explicit guidelines and hierarchical organization
    \item Evolved: An intent-focused template that emphasizes action-oriented concepts and task objectives while maintaining semantic coherence
    \item Model-based: The instagger model \cite{instag}, A template that prioritizes pure intent extraction by focusing on core task objectives and desired outcomes, 
\end{itemize}

Each template was carefully designed to test specific hypotheses about effective instruction encoding, with variations in guidance specificity, semantic preservation requirements, and intent extraction mechanisms (detailed templates in Appendix~\ref{appendix:encode_prompt}).

We assessed their performance along two key dimensions:

\paragraph{Semantic preservation} We measured reconstruction quality using CrossEncoder~\cite{reimers2019sentence}, specifically the \texttt{cross-encoder/stsb-roberta-base} model, to compute semantic similarity between original and reconstructed instructions. We also used ROUGE-L scores as a surface-level textual alignment metric.

\paragraph{Intent extraction capability} We evaluated whether the prompt explicitly requires extracting task intent from instructions, such as identifying action verbs or task objectives. For example, the Enhanced template includes specific guidance like "Focus on action-oriented concepts" and "Hierarchically order by importance."

The reconstruction process used a standardized template (shown in Appendix~\ref{fig:decode_prompt_template}) that provides examples and clear formatting guidelines to ensure consistent instruction regeneration from tags. 

We present three different template designs for tag generation and analyze their performance, with the Evolved template being our I2T prompt shown in Figure~\ref{fig:encode_prompt}:\\

\begin{figure*}[htbp]
\centering
\begin{tcolorbox}[title=Basic Tag Generation Prompt Template (Reconstructed Rate: 0.6536\, Arena Hard: 0.3113)]
\small
Extract exactly three representative tags that capture the core concepts of the following instruction or question. These tags should allow the reconstruction of the original instruction or question while retaining its meaning and intent. Ensure the tags are concise and distinct.\\

[Input]\\
\{instruction\}\\

[Tags]
\end{tcolorbox}
\label{fig:basic_template}
\end{figure*}

\begin{figure*}[htbp]
\centering
\begin{tcolorbox}[title=Enhanced Tag Generation Prompt Template (Reconstructed Rate: 0.5373\, Arena Hard: 0.1992)]
\small
You are a semantic analysis expert. Your task is to extract exactly three essential tags that capture the core semantic concepts of the given instruction or question. The tags should be:
- Concise (1–2 words each)\\
- Hierarchically ordered by importance\\
- Generalizable across similar tasks\\

\#\#\# Guidelines:
1. Focus on action-oriented concepts\\
2. Avoid redundant or overlapping tags\\
3. Use standard terminology when possible\\

\#\#\# Examples:
[Input]\\
In the context of climate change adaptation, analyze how urban planning strategies can be modified to create resilient cities.\\

[Tags]\\
urban resilience, climate adaptation, sustainable development\\

[Input]\\
Design a comprehensive employee training program for a multinational corporation addressing cultural sensitivity.\\

[Tags]\\
corporate training, global workforce, skill development\\

\#\#\# Task:
Given the following instruction, provide exactly three semantic tags following the above format and guidelines:\\

[Input]\\
\{instruction\}\\

[Output Format]\\
tag1, tag2, tag3\\

[Tags]
\end{tcolorbox}
\label{fig:guidelines_template}
\end{figure*}

\begin{figure*}[htbp]
\centering
\begin{tcolorbox}[title=Model-based Tag Annotation Prompt Template (Reconstructed Rate: 0.36\, Arena Hard: 0.1667)]
\small
You are a helpful assistant. Please identify tags of user intentions in the following user query and provide an explanation for each tag. Please respond in the JSON format:

\{ \\
  "tag": "str",\\
  "explanation": "str"\\
\} \\

Query: \{instruction\}\\

Assistant:
\end{tcolorbox}
\label{fig:explanatory_template}
\end{figure*}

\begin{figure*}[htbp]
\centering
\begin{tcolorbox}[title=Decode Prompt Reconstruction Template]
\small
Given the following tags, reconstruct the original instruction as closely as possible.

\#\#\# Example 1:
[Tags]\\
teamwork, collaboration, project management\\

[Instruction]\\
Describe a situation where team collaboration improved the outcome of a project.\\

\#\#\# Example 2:
[Tags]\\
time management, productivity, work-life balance\\

[Instruction]\\
What strategies can be used to improve time management in a busy work environment?\\

\#\#\# Example 3:
[Tags]\\
marketing, strategy, product launch\\

[Instruction]\\
Outline the steps required to create a successful marketing campaign for a new product launch.\\

\#\#\# Your Task:
[Tags]\\
\{tags\}\\

[Instruction]
\end{tcolorbox}
\caption{Decode Prompt Reconstruction Template}
\label{fig:decode_prompt_template}
\end{figure*}

\subsection{Derived Meaning Prompt}\label{appendix:derived_probe}
Below is the complete prompt template used for tag semantic analysis. This template is designed to systematically explore and analyze the semantic dimensions of instruction tags by encouraging divergent thinking followed by convergent analysis. The template guides the model through a structured process of first generating multiple potential interpretations, then carefully consolidating related concepts, and finally distilling the key distinct meanings. This approach helps reveal the semantic richness and utility of different tags in instruction generation: \\

\begin{figure*}[htbp]
\centering
\begin{tcolorbox}[title=Tag Analysis Prompt Template]
\small
For the tag "\{tag\}", follow these steps:

1. Think Different Step: List as many meanings as possible across different domains.\\
2. Merge Step: Reflect on the meanings and combine similar ones into a single, broader concept.\\
3. Final Answer: Provide as few meanings as possible, only listing the most essential and distinct ones.

\#\#\# Examples:

\texttt{[Input]} \\
"video\_share" \\

\texttt{[Think Different Step]}\\
1. A feature to distribute video content\\
2. A social media feature to repost videos\\
3. A platform for users to collaborate on video creation\\
4. A tool for sharing personal video files\\

\texttt{[Merge Step]}\\
- "Distribute video content" and "repost videos" are closely related → merge them into one\\
- "Collaborate on video creation" and "share personal video files" are related but distinct\\

\texttt{[Final Answer]}\\
1. A feature to share videos online\\
That's all\\

\texttt{[Input]}\\
"cloud computing"\\

\texttt{[Think Different Step]}\\
1. A method of delivering computing services over the internet\\
2. A platform for storing and processing data remotely\\
3. A framework for providing software as a service (SaaS) via the internet\\
4. A way for businesses to scale infrastructure without owning physical hardware\\

\texttt{[Merge Step]}\\
- "Delivering computing services" and "storing and processing data remotely" are related → merge into one\\
- "Providing software as a service (SaaS)" is distinct from infrastructure and storage\\
- "Scaling infrastructure without owning hardware" is also distinct\\

\texttt{[Final Answer]}\\
1. A method of delivering computing services and storing data remotely over the internet\\
2. A framework for providing software as a service (SaaS)\\
3. A way for businesses to scale infrastructure without owning physical hardware\\
That's all\\

For the tag "\{tag\}", apply this process and provide the final answer:
\end{tcolorbox}
\caption{Template for analyzing tag semantics through divergent thinking and convergent analysis.}
\label{fig:tag_analysis_template}
\end{figure*}

\subsection{Tag Combination}\label{appendix:tag_combination}

Building on our quantitative control framework through high-utility tags, we introduce tag combination as a principled approach to instruction synthesis. Tag combination merges multiple semantic concepts to create richer, more complex instructions. While prior work has explored combining concepts heuristically, our utility-based selection provides a systematic way to identify and combine high-value tags for optimal instruction generation. The following template (see Figure~\ref{fig:tag_combination_template}) demonstrates our approach to combining tags into coherent instructions.

\begin{figure*}[htbp]
\centering
\begin{tcolorbox}[title=Tag Combination Prompt Template]
\small
Create a comprehensive and challenging task that incorporates these concepts: \{tags\}.\\
Do not explain or provide any additional information. Only return the task/instruction.\\

\#\#\# Response:
\end{tcolorbox}
\caption{Template for combining tags into a task that incorporates their semantic concepts.}
\label{fig:tag_combination_template}
\end{figure*}


\section{Case Study: Tag Examples}\label{appendix:tag_examples}

To further illustrate the relationship between tag value and semantic richness, we present a detailed breakdown of tags across different quartiles (Q1-Q4) based on their Shapley values. Table~\ref{tab:tag_analysis} provides the tag name, occurrence count, average Shapley value, and the number of derived meanings, along with example interpretations for each tag.

\noindent
\begin{minipage}{\textwidth}
    \begin{table}[H]
        \centering
        \small
        \begin{tabular}{|p{0.8cm}|p{3cm}|p{1cm}|p{1.5cm}|p{7cm}|}
            \hline
            \textbf{Quartile} & \textbf{Tag} & \textbf{Count} & \textbf{Avg. Shapley} & \textbf{Derived Meanings} \\
            \hline
            Q4 & data accuracy & 17 & 321.48 & Ensuring correctness; precision; reliability; integrity; completeness \\
            & pytorch & 125 & 321.69 & Deep learning framework; training neural networks; AI research; NLP tool \\
            & payment processing & 48 & 321.72 & Financial transactions; credit card management; mobile payments; cryptocurrency; subscriptions \\
            & food inquiry & 52 & 321.88 & Ingredients and nutrition; restaurant recommendations; availability; allergens; dietary advice \\
            & graph database & 7 & 321.89 & Graph data storage; network analysis; knowledge graphs; recommendation systems \\
            \hline
            Q3 & tableau & 5 & 240.90 & Data visualization tool; dashboard platform \\
            & satire & 7 & 240.90 & Comedy genre; literary exaggeration \\
            & event attendance & 19 & 240.95 & Registration; attendee count; attendee list; event management \\
            & comparative religion & 57 & 240.99 & Religion comparison; texts; rituals; history; ethics; symbols \\
            & personality development & 3 & 241.00 & Self-improvement; emotional intelligence; confidence building; personal growth \\
            \hline
            Q2 & recipe write & 7 & 176.92 & Meal preparation steps; cuisine collection; digital meal planning \\
            & grammar check & 216 & 176.99 & Grammar correctness; punctuation verification; spelling check \\
            & punctuation check & 3 & 177.00 & Punctuation accuracy; error correction; adherence to grammar rules \\
            & physical interaction & 3 & 177.00 & Touch interaction; manual labor; industrial operations; sports activities \\
            & work & 26 & 177.13 & Job; task; workplace; labor \\
            \hline
            Q1 & math expression eval & 3 & 20.13 & Expression evaluation; algebraic solution; simplification \\
            & code execution env & 3 & 24.38 & Code testing; automation scripts; distributed execution \\
            & repayment & 3 & 27.25 & Loan repayment; refunding; restoration \\
            & basic operation & 10 & 28.39 & System operations; core tasks \\
            & number relationship & 3 & 28.52 & Mathematical relations; sequences; dataset correlation \\
            \hline
        \end{tabular}
        \caption{Tag Analysis across Quartiles (Q1-Q4). Higher quartiles indicate higher Shapley values, corresponding to increased semantic richness.}
        \label{tab:tag_analysis}
    \end{table}
\end{minipage}

\section{Case Study: Progressive Enhancement of Web Development Requirements}
This case study demonstrates the progressive enhancement of web development requirements, showing how a simple book listing page evolves into a comprehensive, accessible, and feature-rich web application. The following table presents five iterations of requirements, each building upon the previous version.

\begin{table}[htbp]
\centering
\resizebox{\textwidth}{!}{%
\begin{tabular}{|p{1.5cm}|p{12cm}|p{3cm}|}
\hline
\textbf{Version} & \textbf{Requirements} & \textbf{Tags} \\
\hline
Iter1 & 
\begin{minipage}[t]{12cm}
\vspace{2pt}
Develop an HTML page that dynamically displays a list of books and their authors using CSS for styling. The page should include a search functionality that filters the book list based on the author's name. Additionally, implement a hover effect that reveals a brief description of each book when the user hovers over a book title. Ensure the page is responsive and visually appealing.
\vspace{2pt}
\end{minipage} & 
\begin{minipage}[t]{3cm}
\vspace{2pt}
web\_development, book\_listing, author\_info, css\_styling
\vspace{2pt}
\end{minipage} \\
\hline
Iter2 & 
\begin{minipage}[t]{12cm}
\vspace{2pt}
Develop an HTML page that dynamically displays a list of books and their authors using CSS for styling. The page should include a search functionality that filters the book list based on the author's name, as well as a pagination system to handle a large number of books. Additionally, implement a hover effect that reveals a brief description of each book when the user hovers over a book title. Ensure the page is responsive and visually appealing, and optimize the user experience by incorporating smooth transitions and animations for the hover effect and pagination.
\vspace{2pt}
\end{minipage} & 
\begin{minipage}[t]{3cm}
\vspace{2pt}
web\_development, user\_interface, responsive\_design, user\_experience
\vspace{2pt}
\end{minipage} \\
\hline
Iter3 & 
\begin{minipage}[t]{12cm}
\vspace{2pt}
Develop an HTML page that dynamically displays a list of books and their authors using CSS for styling. The page should include a search functionality that filters the book list based on the author's name, as well as a pagination system to handle a large number of books. Additionally, implement a hover effect that reveals a brief description of each book when the user hovers over a book title. Ensure the page is responsive and visually appealing, and optimize the user experience by incorporating smooth transitions and animations for the hover effect and pagination. Furthermore, enhance the accessibility of the page by ensuring that all interactive elements are keyboard navigable and that the page is compatible with screen readers. Consider the use of ARIA roles and properties to improve the accessibility of the search functionality and pagination system.
\vspace{2pt}
\end{minipage} & 
\begin{minipage}[t]{3cm}
\vspace{2pt}
web\_development, user\_interface, responsive\_design, accessibility
\vspace{2pt}
\end{minipage} \\
\hline
Iter4 & 
\begin{minipage}[t]{12cm}
\vspace{2pt}
Develop an HTML page that dynamically displays a list of books and their authors using CSS for styling. The page should include a search functionality that filters the book list based on the author's name, as well as a pagination system to handle a large number of books. Additionally, implement a hover effect that reveals a brief description of each book when the user hovers over a book title. Ensure the page is responsive and visually appealing, and optimize the user experience by incorporating smooth transitions and animations for the hover effect and pagination. Furthermore, enhance the accessibility of the page by ensuring that all interactive elements are keyboard navigable and that the page is compatible with screen readers. Consider the use of ARIA roles and properties to improve the accessibility of the search functionality and pagination system. Additionally, implement a feature that allows users to sort the book list by title, author, or publication year. Ensure that the sorting functionality is accessible and provides clear visual feedback to users. Finally, optimize the page's performance by implementing lazy loading for images and minimizing the use of heavy CSS animations to ensure a smooth user experience on devices with limited processing power.
\vspace{2pt}
\end{minipage} & 
\begin{minipage}[t]{3cm}
\vspace{2pt}
web\_development, user\_experience, accessibility, performance\_optimization
\vspace{2pt}
\end{minipage} \\
\hline
\end{tabular}%
}
\caption{Case Study of Progressive Enhancement of Web Development Requirements}
\label{tab:web_dev_requirements}
\end{table}

\end{document}